\documentclass{article}
\usepackage{verbatim}

\usepackage[utf8]{inputenc} 
\usepackage[T1]{fontenc}    
\usepackage{hyperref}       
\usepackage{url}            
\usepackage{booktabs}       
\usepackage{amsfonts}       
\usepackage{nicefrac}       
\usepackage{microtype}      
\usepackage{xcolor}         
\usepackage{bbm}
\usepackage{amsmath}
\usepackage{animate}
\usepackage{media9}
\usepackage[numbers]{natbib}
\usepackage{amsmath, graphicx,amssymb, hyperref, bbm, multirow, listings}
\usepackage{verbatim} 
\usepackage{tikz}
\usepackage{todonotes}
\usepackage[affil-it]{authblk}
\usepackage[left=1in, right=1in, top=1in, bottom=1in]{geometry}
\usepackage{color}
\usepackage{newfloat}
\DeclareFloatingEnvironment[
    fileext=los,
    listname={List of Supplementary Figures},
    name=Supplementary Figure,
    placement=tbhp,
    within=none,
]{suppfigure}
\DeclareFloatingEnvironment[
    fileext=lot,
    listname={List of Supplementary Tables},
    name=Supplementary Table,
    placement=tbhp,
    within=none,
]{supptable}
\newcommand{\beginsupplement}{%
    \setcounter{table}{0}
    \setcounter{figure}{0}
    \renewcommand{\thetable}{S\arabic{table}}
    \renewcommand{\thefigure}{S\arabic{figure}}
}

\title{Residual Corrective Diffusion Modeling for Km-scale Atmospheric Downscaling}

\author[1,*,a]{Morteza Mardani}
\author[1,*]{Noah Brenowitz}
\author[1,*]{Yair Cohen}
\author[1]{Jaideep Pathak}
\author[1]{Chieh-Yu Chen}
\author[2]{Cheng-Chin Liu}
\author[1]{Arash Vahdat}
\author[1]{Mohammad Amin Nabian}
\author[1]{Tao Ge}
\author[1]{Akshay Subramaniam}
\author[1]{Karthik Kashinath}
\author[1]{Jan Kautz}
\author[1]{Mike Pritchard}

\affil[1]{NVIDIA, Santa Clara, CA. 95050, USA}
\affil[2]{Central Weather Administration, 64, Gongyuan Road, Taipei 100006, Taiwan}
\affil[*]{These authors contributed equally}
\affil[a]{Corresponding author: \texttt{mmardani@nvidia.com}}

\begin{document}

\maketitle

\begin{abstract}
The state of the art for physical hazard prediction from weather and climate requires expensive km-scale numerical simulations driven by coarser resolution global inputs. Here, a generative diffusion architecture is explored for downscaling such global inputs to km-scale, as a cost-effective machine learning alternative. The model is trained to predict 2km data from a regional weather model over Taiwan, conditioned on a 25km global reanalysis. To address the large resolution ratio, different physics involved at different scales and prediction of channels beyond those in the input data, we employ a two-step approach where a UNet predicts the mean and a corrector diffusion (CorrDiff) model predicts the residual. CorrDiff exhibits encouraging skill in bulk MAE and CRPS scores. The predicted spectra and distributions from CorrDiff faithfully recover important power law relationships in the target data. Case studies of coherent weather phenomena show that CorrDiff can help sharpen wind and temperature gradients that co-locate with intense rainfall in cold front, and can help intensify typhoons and synthesize rain band structures. Calibration of model uncertainty remains challenging. The prospect of unifying methods like CorrDiff with coarser resolution global weather models implies a potential for global-to-regional multi-scale machine learning simulation.
\end{abstract}

\section{Introduction}
Coarse-resolution 25-km global weather prediction is undergoing a machine learning renaissance with the recent advance of autoregressive machine learning models trained on global reanalysis \cite{benbouallegue2023rise, pathak2022fourcastnet, bonev2023spherical, bi2023accurate, lam2022graphcast,croitoru2023diffusion, li2023fuxi, price2023gencast, bodnar2024aurora, lang2024aifs}. However, many applications of weather and climate data require kilometer-scale forecasts: e.g., risk assessment and capturing local effects of topography and human land use ~\cite{gutowski2020ongoing}. Globally, applying ML at km-scale resolution poses significant challenges since training costs are superlinear with respect to the resolution of training data. Moreover, predictions from global km-scale physical simulators are not yet well tuned, so available training data can have worse systematic biases than coarse-resolution or established regional simulations \cite{stevens2019dyamond,hohenegger2023icon}, and current data tends to cover short periods of time. Such datasets are also massive, difficult to transfer between data centers and frequently not produced on machines attached to significant AI computing resources like GPUs. 

In contrast, for regional simulation, using ML to conditionally generate km-scales is attractive. High-quality training data are available as many national weather agencies couple km-scale numerical weather models in a limited domain to coarser resolution global models \cite{dutta2019regional}  -- a process called dynamical downscaling. Since these predictions are augmented by data assimilation from ground-based precipitation radar and other sensors, good estimate of regional km-scale atmospheric states exists \cite{chen2020improving}. Such dynamical downscaling is computationally expensive, which limits the number of ensemble members used to quantify uncertainties \cite{nishant2023comparison}. 

A common inexpensive alternative is to learn a statistical downscaling from these dynamical downscaling simulations and observations \cite{wilby1998statistical}. This is typically done by learning the values of several parameters of a statistical mapping (e.g. quantile mapping, generalized linear regression) that best match a regional high resolution  data \cite{bano2020configuration}. In this context, ML downscaling enters as an advanced (non linear) form of statistical downscaling \cite{rampal2024enhancing} with potential to exceed the fidelity of conventional statistical downscaling. 

Several ML methods have been previously used for downscaling \cite{bischoff2023unpaired, sebbar2023machine, geiss2020radar, teufel2023physics, nishant2023comparison, vosper2023deep, adewoyin2021tru}. Convolutional Neural Networks have shown promise in globally downscaling climate (100km) data to weather scales (25km) \cite{mu2020climate, rodrigues2018deepdownscale, bano2020configuration, rampal2022high}. However, such deterministic ML approaches require interventions to produce useful probabilistic results, such as ensemble inference \cite{rodrigues2018deepdownscale} or predicting the parameters of an assumed distribution \cite{bano2020configuration}).

The stochastic nature of atmospheric physics at km-scale \cite{selz2015upscale} renders downscaling inherently probabilistic, making it natural to explore generative models at these scales. Generative Adversarial Networks (GANs) have been tested, including for forecasting precipitation at km-scale in various regions \cite{leinonen2020stochastic, price2022increasing, harris2022generative, ravuri2021skilful, gong2023enhancing, vosper2023deep}; see the latter for a good review. Training GANs, however, poses several practical challenges including mode collapse, training instabilities, and difficulties in capturing long tails of distributions \cite{xiao2022DDGAN, kodali2017convergence, salimans2016improved}.

Alternatively, diffusion models offer training stability \cite{ho2020denoising, dhariwal2021diffusion} alongside demonstrable skill in probabilistically generating km-scales.  
\cite{addison2022machine} used a diffusion model for predicting rain density in the UK from vorticity as an input, thus demonstrating potential for channel synthesis. \cite{hatanaka2023diffusion} used a diffusion model for downscaling solar irradiance in Hawaii with a 1 day lead time, demonstrating the ability to simultaneously forecast. Moreover, diffusion models have been used directly for probabilistic weather forecasting and nowcasting \cite{leinonen2023latent, li2023seeds, nath2023forecasting, stock2024diffobs}  -- including global ensemble predictions that outperform conventional weather prediction on a range of important stochastic metrics at 0.25-degree resolution \cite{price2023gencast}. See table \ref{tab:related_works} in \ref{related_works} for more details.

Building upon these works, we turn to our challenge of interest -- stochastically downscaling multiple variables simultaneously while also transferring input information to predict a new field (i.e., channel synthesis). If successful, this paves the way towards ML downscaling systems that produce regional high-resolution weather as a postprocessing of coarser global predictions. As a proof of concept we will demonstrate such a ML model trained for the region surrounding Taiwan.

Details follow. The key contributions of this paper are:

\begin{enumerate}
    \item A physics-inspired, two-step approach (CorrDiff) to simultaneously learn mappings between low- and high-resolution weather data across multiple variables with high fidelity alongside new channel synthesis. 
    \item For the case studies considered, CorrDiff adds physically realistic improvements to the representation of under-resolved coherent weather phenomena -- frontal systems and typhoons.
    \item CorrDiff is sample-efficient, learning effectively from just 3 years of data.
    \item CorrDiff on a single GPU is at least 22 times faster and 1,300 times more energy efficient than the numerical model used to produce its high-resolution  training data, which is run on 928 CPU cores, see \ref{sec:speedup} for details.
\end{enumerate}

\section{Generative downscaling: Corrector diffusion model}
\label{sec:generativedownscaling}
Consider a specific region on Earth, mapped onto a two-dimensional grid. Our input ${\bf y} \in \mathbbm{R}^{c_{\mathrm{in}} \times m \times n}$ is a low-resolution meteorological data taken from a 25-km global reanalysis, or weather forecasting model (e.g., FourCastNet \cite{pathak2022fourcastnet, lam2022graphcast, bi2023accurate}, or the Global Forecast System (GFS) \cite{ncep_rda_ucar2015}). Here, \(c_\mathrm{in}\) represents the number of input channels and $m,n$ represent the dimensions of a 2D subset of the globe. Our targets ${\bf x} \in \mathbbm{R}^{c_\mathrm{out} \times p \times q}$ come from corresponding data aligned in time \(c_\mathrm{out}\) but having higher resolution, i.e., $p > m$ and $q > n$. 

In our proof of concept we use the ERA5 reanalysis as input, over a subregion surrounding Taiwan, with $m=n=36$, $c_{in}=12$ and $c_{out}=4$. See Table \ref{tab:input_output} for details about the inputs and outputs. The target data are 12.5 times higher resolution ($p=q=448$) and were produced using a radar-assimilating Weather Research and Forecasting (WRF) physical simulator \cite{powers2017weather} provided by the Central Weather Administration of Taiwan (CWA) \cite{chen2020improving} (i.e., CWA-WRF), which employs a dynamical downscaling approach. Though imperfect, WRF is a SOTA model for km-scale weather simulations and is used operationally by several national weather agencies. 

The goal of probabilistic downscaling is to mimic the conditional probability density $p({\bf x}|{\bf y})$. To learn $p({\bf x}|{\bf y})$ we employ a diffusion model. Such models learn stochastic differential equations (SDEs hereafter) through the concept of score matching \cite{ho2020denoising, song2019generative, karras2022elucidating, song2021denoising,batzolis2021conditional}, with a forward and a backward processes working in tandem. In the forward process, noise is gradually added to the target data until the signal becomes indistinguishable from noise. 

The backward process then involves denoising the samples using a dedicated neural network to eliminate the noise. Through this sequential denoising process, the model iteratively refines the samples, bringing them closer to the target data distribution. The denoising neural network plays a critical role in this convergence, providing the necessary guidance to steer the samples towards accurate representations of the original data.

The development of CorrDiff was motivated by the limitations observed when using conditional diffusion models to directly learn \( p(\mathbf{x}|\mathbf{y}) \). This approach showed slow convergence and resulted in poor-quality images with incoherent structures. This was surprising because conditional diffusion models have been successfully applied to super-resolution tasks in natural image restoration, as demonstrated in works like \cite{rombach2022high}. We hypothesize that the significant distribution shift between the input variables and challenging target variables, particularly the 1-hour maximum derived radar reflectivity (hereafter referred to as radar reflectivity), necessitates high noise levels during the forward process and numerous steps in the backward process. Our experiments indicated that these requirements hindered learning and compromised sample fidelity \cite{song2020improved}. This issue is particularly relevant for the downscaling task, which must account for large spatial shifts, correct biases in static features like topography, and synthesize entirely new channels like radar reflectivity. By comparison, the task of super-resolution in natural images is much simpler, as it typically involves local interpolation and does not face the same level of distributional challenges.

\begin{figure}
\centering
\hspace{0mm}\includegraphics[width=0.9\textwidth]{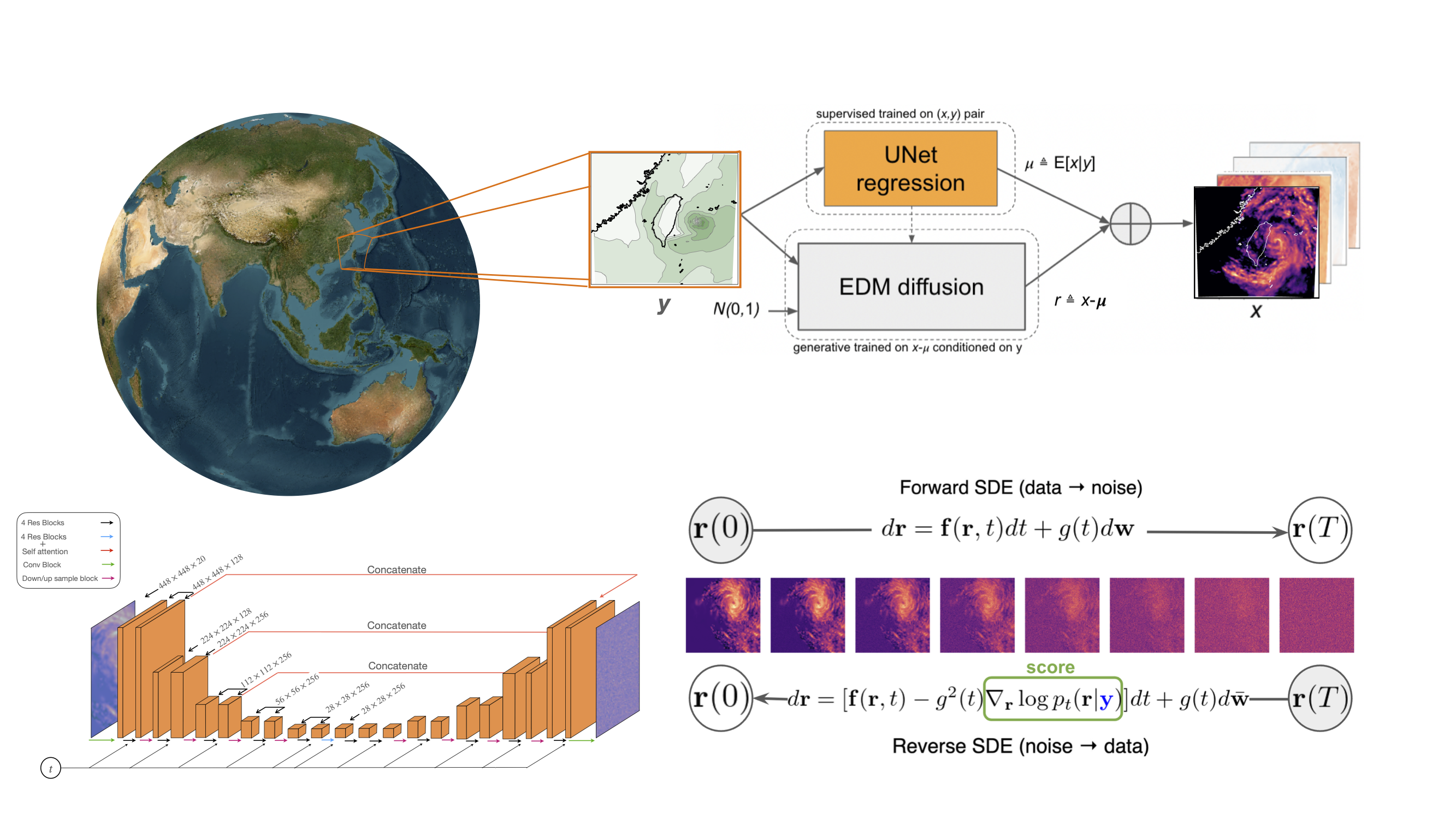}  
\caption{The workflow for training and sampling CorrDiff for generative downscaling. Top: Coarse-resolution global weather data at 25 km scale is used to first predict the mean $\boldsymbol{\mu}$ using a regression model, which is then stochastically corrected using an Elucidated Diffusion Model (EDM) ${\bf r}$, together producing the probabilistic high-resolution 2 km-scale regional forecast. Bottom right: diffusion model is conditioned with the coarse-resolution input to generate the residual ${\bf r}$ after a few denoising steps. Bottom left: the score function for diffusion is learned based on the UNet architecture.}  \label{fig:CorrDiff_diagram}
\end{figure}

To sidestep these challenges, we decompose the generation into two steps (Fig. \ref{fig:CorrDiff_diagram}). The first step predicts the conditional mean using (UNet) regression (see also \ref{sec:architecture-description} and \ref{fig:UNet} for details), and the second step learns a correction using a diffusion model as follows:
\begin{align}
        {\bf x} = \underbrace{\mathbbm{E}[{\bf x} | {\bf y}]}_{:={\boldsymbol{\mu}} 
 ({\rm regression})} + \quad \underbrace{({\bf x} - \mathbbm{E}[{\bf x} | {\bf y}])}_{:= {\bf r}  (\rm generation)}, \label{eq:decomposition}
\end{align}
where ${\bf y}$ and ${\bf x}$ are the input and target respectively. This signal decomposition is inspired by Reynolds decomposition in fluid-dynamics \cite{pope2000turbulent} and climate data analytics. Assuming the regression learns the conditional mean accurately, i.e., ${\bf \mu} \approx \mathbbm{E} [{\bf x} | {\bf y}]$, the residual is zero mean, namely $\mathbbm{E} [{\bf r} | {\bf y}] \approx 0$, and as a result ${\rm var}({\bf r} | {\bf y}) = {\rm var}({\bf x} | {\bf y})$. Accordingly, based on the {\it law of total variance} \cite{blitzstein2019introduction}, one can decompose the variance as
\begin{align}
    {\rm var}({\bf r}) & = \mathbbm{E} \big[ {\rm var}({\bf r} | {\bf y})  \big] +  \underbrace{{\rm var} \big( \mathbbm{E} [{\bf r} | {\bf y}] \big)}_{=0} 
    \leq \mathbbm{E} \big[{\rm var}({\bf x} | {\bf y}) \big] + \underbrace{ {\rm var} \big(\mathbbm{E} [{\bf x}| {\bf y}]\big)}_{ \geq 0}   = {\rm var}({\bf x}).   \label{eq:var_decompos_res}
\end{align}
That is, the residual formulation reduces the variance of the target distribution. According to \eqref{eq:var_decompos_res}, the variance reduction is more pronounced when ${\rm var}(\mathbbm{E} [{\bf x}| {\bf y}])$ is large, e.g., in the case of typhoons. For our specific target data we find that the actual variance reduction is significant, especially at large scales; see section \ref{sec:acf} and Figure \ref{fig:acf}.
To sum it up, the main idea of CorrDiff is that learning the distribution $p(\bf r)$ can be much easier than learning the distribution $p(\bf x)$. Since modeling multi-scale interactions is a daunting task in many physics domains, we expect this approach could be widely applied. More details are described in Section~\ref{sec:methods} and the outline is depicted in Fig.~\ref{fig:CorrDiff_diagram}.

\def\figwidth{4in}
\begin{figure}
    \centering
    \includegraphics[width=0.7\textwidth, clip]{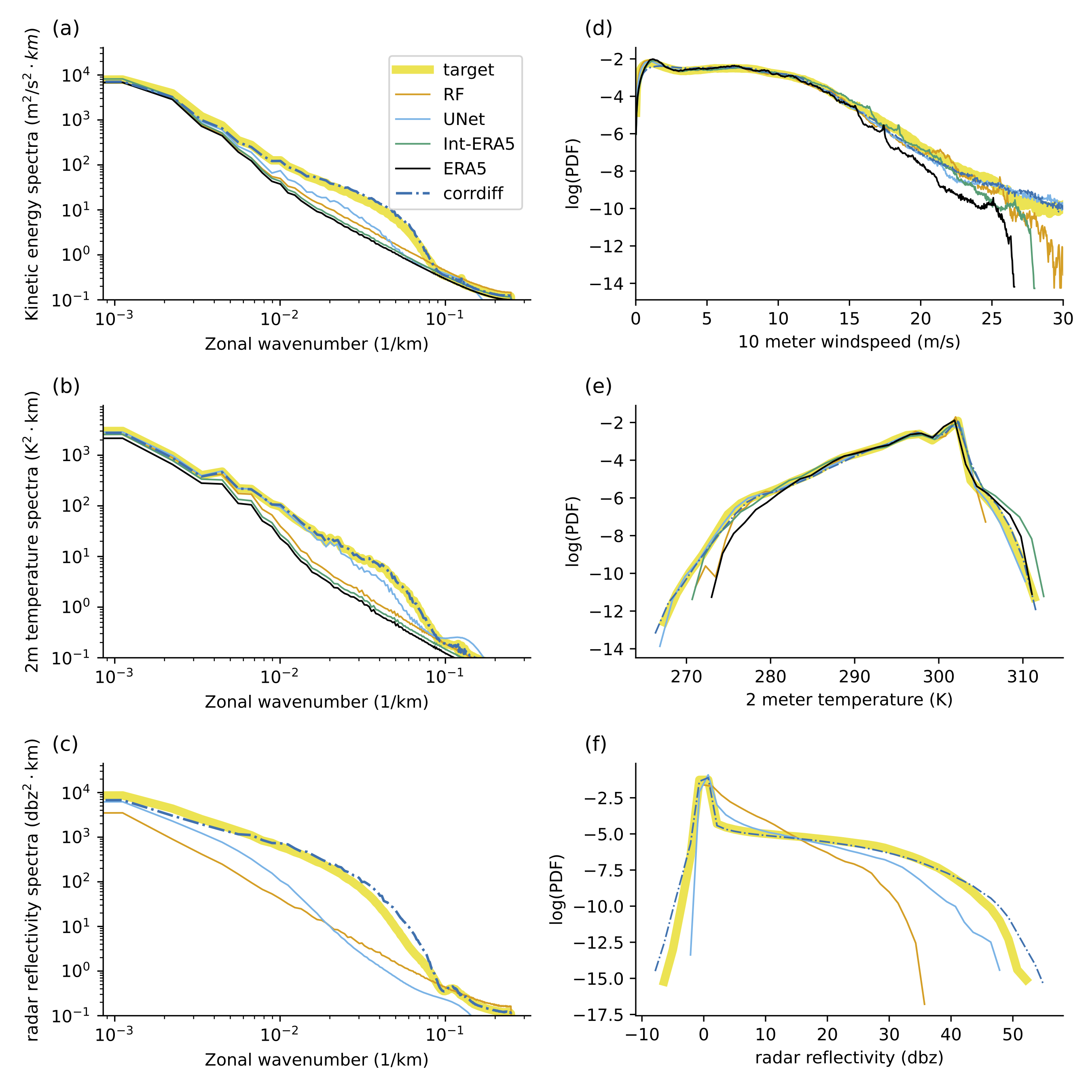}
    \caption{Power spectra and distributions for the interpolated ERA5 input, CorrDiff, RF, UNet, and WRF. These results reflect reductions over space, time and for \textit{CorrDiff} across 32 different samples per each time. Left: Power spectra for kinetic energy (top), 2-meter temperature (middle) and radar reflectivity (bottom). Right: distributions of windspeed, (top), 2-meter temperature (middle) and radar reflectivity (bottom). Radar reflectivity is not included in the ERA5 dataset. We show the log-PDF to highlight the differences at the tails of the distributions. Here wavenumber is the inverse of a wavelength.}
    \label{fig:spectra_distributions}
\end{figure}

Our target (WRF) dataset spans 2018 through 2021 at hourly time resolution. We use 2018 through 2020 for training and the rest for testing. We additionally use several days of typhoon data from 2023 and some snapshots of a coherent frontal weather system from 2022 for testing case studies. The input (coarse resolution) data are taken from the ERA5 reanalysis for the corresponding times. The UNet and a random forest are used as baselines. See Section \ref{sec:methods} and Table \ref{tab:input_output} for details. 

\section{Results}

In this section CorrDiff downscaling is compared with the input and target data as well as with several baseline models. A common set of 205 randomly selected out-of-sample date and time combinations from 2021 is used for computing metrics and spectra and for intercomparing CorrDiff with the baseline models. For CorrDiff ensemble predictions are examined using a 32-member ensemble; larger ensembles do not meaningfully modify the key findings below (not shown). 

\subsection{Baseline Models}
 As baselines, we use an interpolation of the condition data (ERA5), a Random Forest (RF) and the regression step of CorrDiff (UNet).
Using the same 12 low-resolution input channels we fit an RF for each of the 4 output channels with 100 trees and the default hyperparameters.
The RF is applied independently at each horizontal location similar to a $1\times1$ convolution.
While crude, this RF provides a simple (and easily tuned) baseline for the performance of the UNet. To ensure the best performance for each channel individually, we train separate RFs for each output channel.

\subsection{Skill}
When comparing the CRPS of CorrDiff with the MAE of the UNet and the other baselines, CorrDiff exhibits the most skill, followed by the UNet, the random forest (RF) and the interpolation of ERA5 (Table (\ref{tab:big-score}). The slight degradation in MAE of CorrDiff compared to that of the UNet is expected, as the diffusion model optimizes the Kullback–Leibler divergence as opposed to optimizing for MAE loss optimized by the UNet (see Section \ref{sec:Denoising}). In table \ref{tab:pooled_scores} we show that pooled scores (CRPS and MAE) tell a similar story.

\begin{table}
\centering
\begin{tabular}{@{}llrrrr@{}}
\begin{tabular}{lrrrr}
\toprule
 & Radar & t2m & u10m & v10m \\
\midrule
CorrDiff (CRPS) & 1.90 & 0.55 & 0.86 & 0.95 \\
CorrDiff (MAE) & 2.54 & 0.65 & 1.08 & 1.19 \\
UNet & 2.51 & 0.64 & 1.10 & 1.21 \\
RF & 3.56 & 0.81 & 1.14 & 1.26 \\
ERA5 & - & 0.97 & 1.17 & 1.27 \\
\bottomrule
\end{tabular}
\end{tabular}
\vspace{2mm}
\caption{MAE and CRPS scores evaluated from 205 date and time combinations taken randomly from the out-of-sample year (2021). For CorrDiff the CRPS was computed using 32 ensemble members and the MAE is computed for the ensemble mean. For deterministic predictions given by all other models, MAE and CRPS are equivalent. The differences between CorrDiff, UNet, and RF are all statistically significant (see SI Section \ref{sec:significance}). CorrDiff has lower CRPS than the UNet in 205/205 of the validation times.}
\label{tab:big-score}
\end{table}

\subsection{Spectra and distributions}
Relative to deterministic baselines, CorrDiff significantly improves the realism of power spectra for 10-meter kinetic energy (KE), 2-meter temperature and synthesized radar reflectivity. Variance missing from the UNet is restored by the corrective diffusion (blue-dashed vs blue-solid) -- especially for the radar reflectivity channel at all length scales (Fig.~\ref{fig:spectra_distributions}c), but also for kinetic energy between 10--200 km length scales (Fig.~\ref{fig:spectra_distributions}a) and to a lesser extent for temperature on 10--50 km length scales. Temperature downscaling is an easier task that is expected to be mostly driven by sub-grid variations in topography that can be learned deterministically from the static grid embeddings. Evidently, synthesizing radar from only indirectly related inputs is the task that most benefits from the corrective diffusion component of CorrDiff. 

This is corroborated by analysis of probability distributions (Fig.~\ref{fig:spectra_distributions}d-f) -- for the radar reflectivity channel both the UNet and RF fail to produce realistic statistics, but CorrDiff is able to match the target distribution between 0 and 43 dbz while significantly improving on the UNet (Fig.~\ref{fig:spectra_distributions}f). In contrast to the radar channel, the hot and cold tails of the CorrDiff-generated surface temperature distribution are only incrementally improved relative to the UNet (Fig.~\ref{fig:spectra_distributions}e) and the overall windspeed PDF is virtually unchanged relative to the UNet, despite the scale-selective variance enhancements noted previously. Overall, CorrDiff produces encouraging probability distributions, with the caveat that apparent agreement of generated tail structures should be viewed as provisional given that our chosen validation sample of 205 independent calendar times imperfectly samples especially low likelihood/high-impact extremes.

While encouraging, CorrDiff's emulation of radar statistics is also imperfect; generated radar variance is somewhat under-estimated on length scales greater than 100 km and over-estimated for the 10--50 km length scales (Fig.~\ref{fig:spectra_distributions}c), associated with an overall overdispersive PDF (Fig.~\ref{fig:spectra_distributions}f).

\subsection{Model Calibration}
\label{sec:calibration}
\begin{figure}
    \centering
    \includegraphics[width=0.4\textwidth]{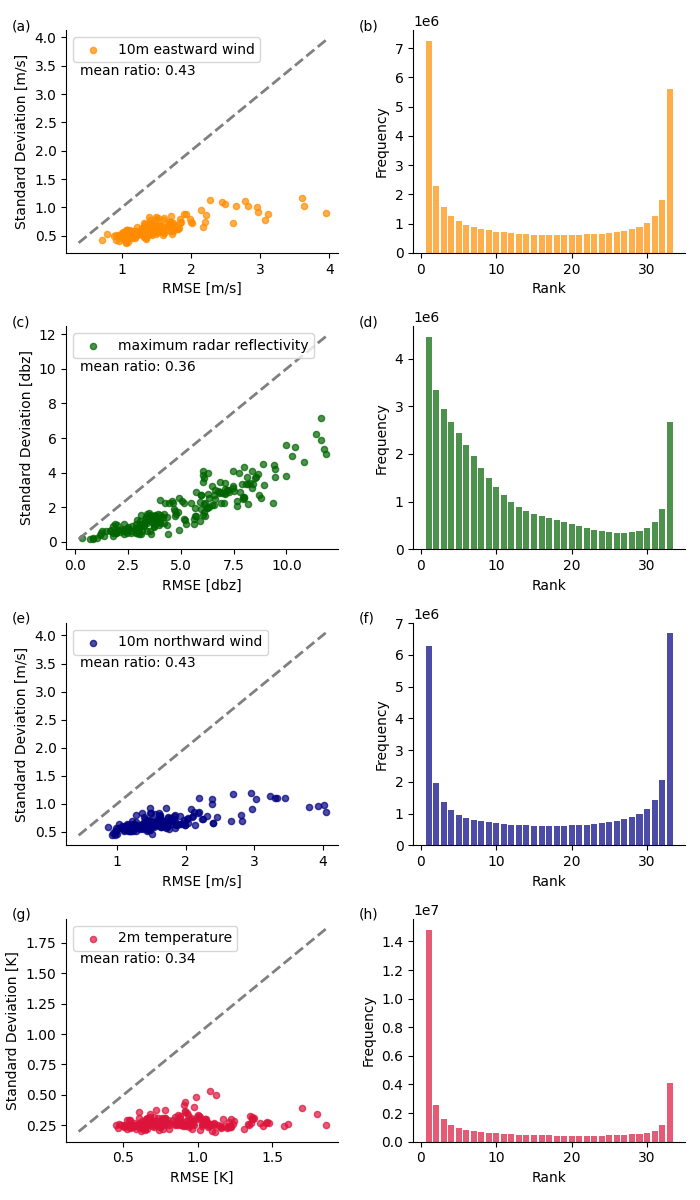}
    \caption{Evaluation of model calibration base do the same validation set used in figure \ref{fig:spectra_distributions} and \ref{tab:big-score}. Left column - the ensemble standard deviation as a function of the RSME of mean prediction for the 4 channels. The standard deviation is adjusted with a factor $\sqrt(1+1/n)$ so that a ratio of one represents a perfectly tuned model. Right column shows the corresponding rank histograms.}
    \label{fig:calibration}
\end{figure}

Analysis of the ensemble spread of our 32-member CorrDiff predictions shows they are not yet optimally calibrated. Figure~\ref{fig:calibration} demonstrates that the predictions are overall under-dispersive for most channels -- the ensemble spread is too small relative to ensemble mean error and rank histograms indicate that observed values frequently fall above or below the range of predicted values. Optimizing the stochastic calibration of CorrDiff is a logical area for future development. 

\subsection{Case studies: downscaling coherent structures}

\begin{figure}
    \centering
    \includegraphics[width=1.0\textwidth,trim={50pt 35pt 50pt 35pt},clip]{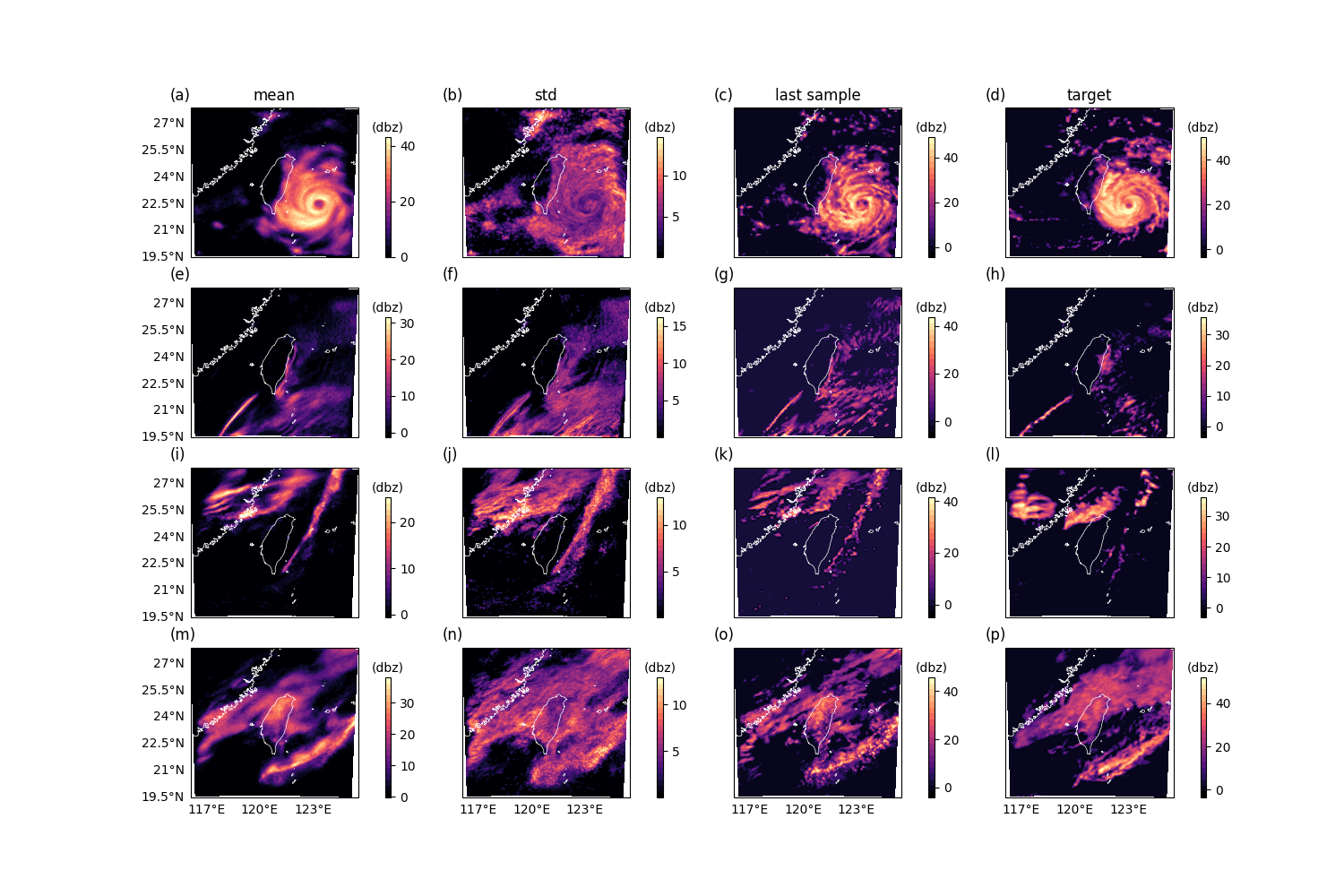}
    \caption{Demonstration of the stochastic prediction of radar reflectivity (in dBZ). Top to bottom: 2023-09-03 00:00:00 , 2021-02-17 21:00:00, 2021-03-04 01:00:00 and 2022-02-13 20:00:00 UTC. Left to right: sample mean, sample standard deviation, sample number 32 and the target forecast.}
    \label{fig:radar_ref}
\end{figure}

 We now turn our attention to specific weather regimes, which are important to examine since aggregate skill scores and spectra can be more easily gamed and mask symptoms of spatial incoherence. Fig.~\ref{fig:radar_ref} illustrates the variability of the generated radar reflectivity field on four separate dates corresponding to distinct Taiwanese weather events. Two dates are chosen randomly (Fig.~\ref{fig:radar_ref} e-k). The other two correspond to a dates where coherent events such as typhoon (Fig.~\ref{fig:radar_ref} a-d) and frontal event (Fig.~\ref{fig:radar_ref} m-o) are present; these dates are further analyzed in the following sections and more examples of both of these phenomena are provided in the Appendix for additional context \ref{section:additional_cases}. The standard deviation across our ensemble of 32 generated CorrDiff samples (second column from the left) is roughly 20\% of the magnitude of the ensemble mean (left column). The CorrDiff prediction for an arbitrary ensemble member (last sample; 32nd member; third column from the left) is useful to compare to the target data (right column). However, due to the stochastic nature of the generation, some disagreement in detailed patterns and positioning should be expected. The similarity between the first and the third columns highlights the role of the mean UNet prediction in forming large-scale coherent structures, such as the positioning of rainbands within typhoon Haikui (2023), top row, and frontal systems, bottom row. The additional fine-scale structure reflecting the stochastic physics contributed by the diffusion model is seen in the third column of Fig.~\ref{fig:radar_ref}. Further comparison across independent generated samples is presented in an animation in \ref{fig:animation} in \ref{sample_diversity} that is helpful for appreciating the portion of the generated image that is governed by the corrective diffusion subcomponent of CorrDiff.

\subsubsection{Frontal system case study}

Frontal systems are an example of organized atmospheric systems. A cold front is a sharp change in temperature and winds associated with a mature, mid-latitude, cyclonic storm. As the front moves eastward, the cold air pushes the warm air to its east upward. This upward motion leads to cooling, condensation and ultimately rainfall. That is, these physics should manifest as multi-variate relationships with linked fine scale structures of two wind vector components and temperature that should co-locate with radar reflectivity. 

\begin{figure}
    \centering
    \includegraphics[width=1.0\textwidth]{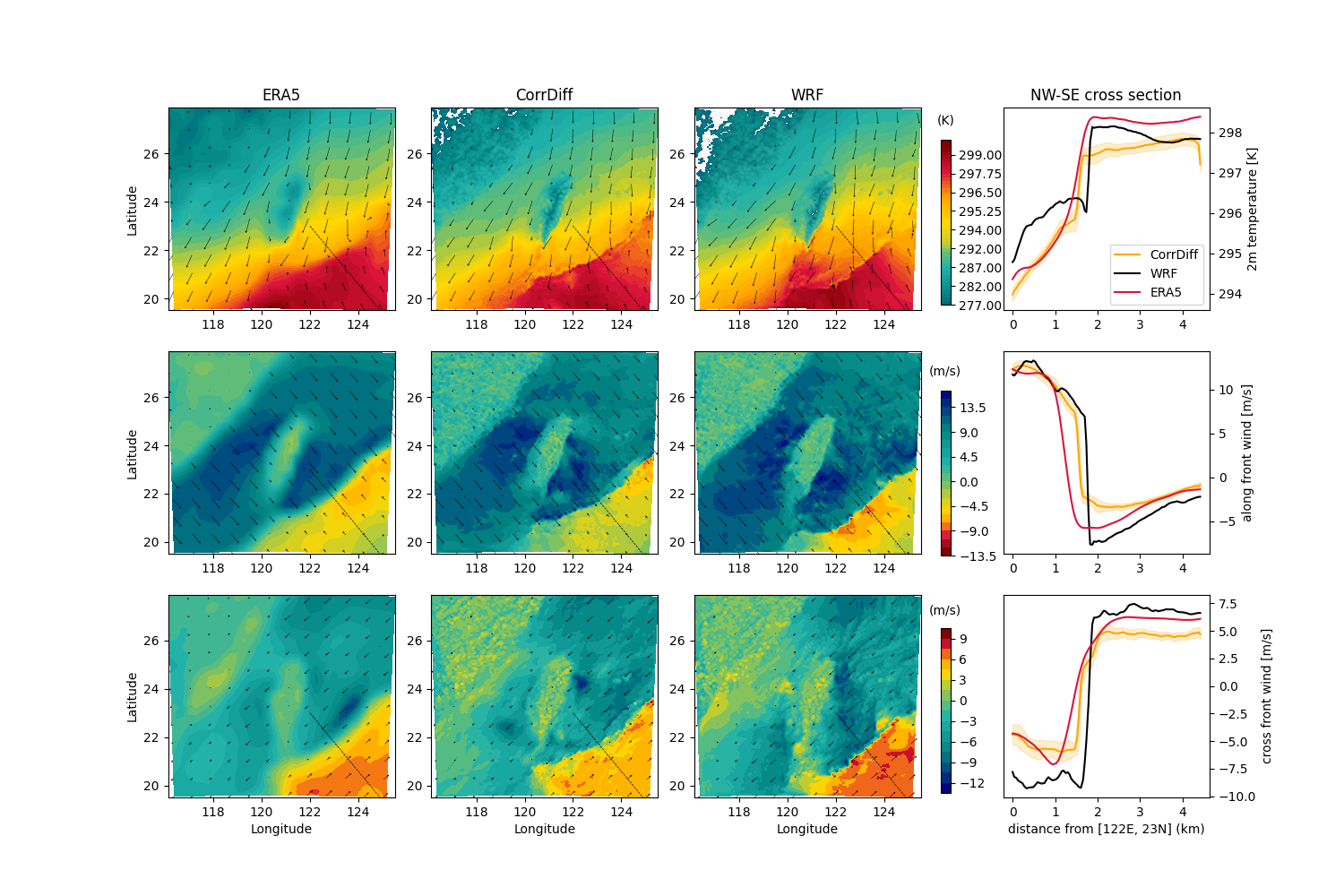}
    \caption{Examining the downscaling of a cold front on 2022-02-13 20:00:00 UTC. Left to right: prediction of ERA5, CorrDiff and Target for different fields, followed by their averaged cross section from 21 lines parallel to the thin dashed line in the contour figures. Top to bottom: 2-meter temperature (arrows are wind vectors), along front wind (arrows are along-front component of the wind vector) and across front wind (arrows are across-front component of the wind vector). At the right column the cross sections of the WRF (black line) and ERA5 (red line) are compared with the mean of a 32 member ensemble prediction from CorrDiff (orange line) where the shading shows $\pm$ one standard deviation.}
    \label{fig:generation-front}
\end{figure}

Fig.~\ref{fig:generation-front} shows an example of CorrDiff downscaling a cold front. Examining the target data (WRF in third column), the position of the front is clearly visible in the southeast portion of the domain, where a strong horizontal 2-meter temperature gradient (top) co-locates with both a strong divergence of the across-front wind (bottom) and a reversal in direction of the along-front wind on either side of the temperature front (middle). Compared to the target data the ERA5 representation of this front is smoother. CorrDiff partially restores sharpness to the front by increasing the horizontal gradients across all three field variables. Although the generated front has some differences in morphology compared to the ground truth, the consistency of its morphology across winds and temperature is reassuring. The intense rainfall associated with the convergence at the front can be seen in the radar reflectivity for the same date and time in bottom row of Fig.~\ref{fig:radar_ref}. The generated radar reflectivity is appropriately concentrated near the sharpened frontal boundary at the cold sector. We expand this analysis of the frontal boundaries across more samples in Section \ref{section:additional_front}, which reveals that CorrDiff consistently adjusts the winds, temperature and  radar reflectivity at the front, but that its skill in sharpening frontal gradients exhibits case-to-case variability.

\subsubsection{Tropical Cyclone case study}
\label{sec:hurricane}

Downscaling typhoons (i.e., tropical cyclones) is especially complicated, helpfully revealing the limitations of CorrDiff for representing extreme events. Not only are typhoons extremely rare in our training data, but the average radius of maximum winds of a tropical cyclone is less than 100km such that at the 25km resolution of our input data tropical cyclones are only partially resolved. This leads to a cyclonic structures that are too wide in horizontal extent and too weak in wind intensity compared with high-resolution  models or observations \cite{bian2021well}. A useful downscaling model should simultaneously correct their size and intensity in addition to generating appropriate fine-scale structure. 

\begin{figure}
    \centering
        \hspace{0mm}\includegraphics[width=0.8\textwidth,clip]{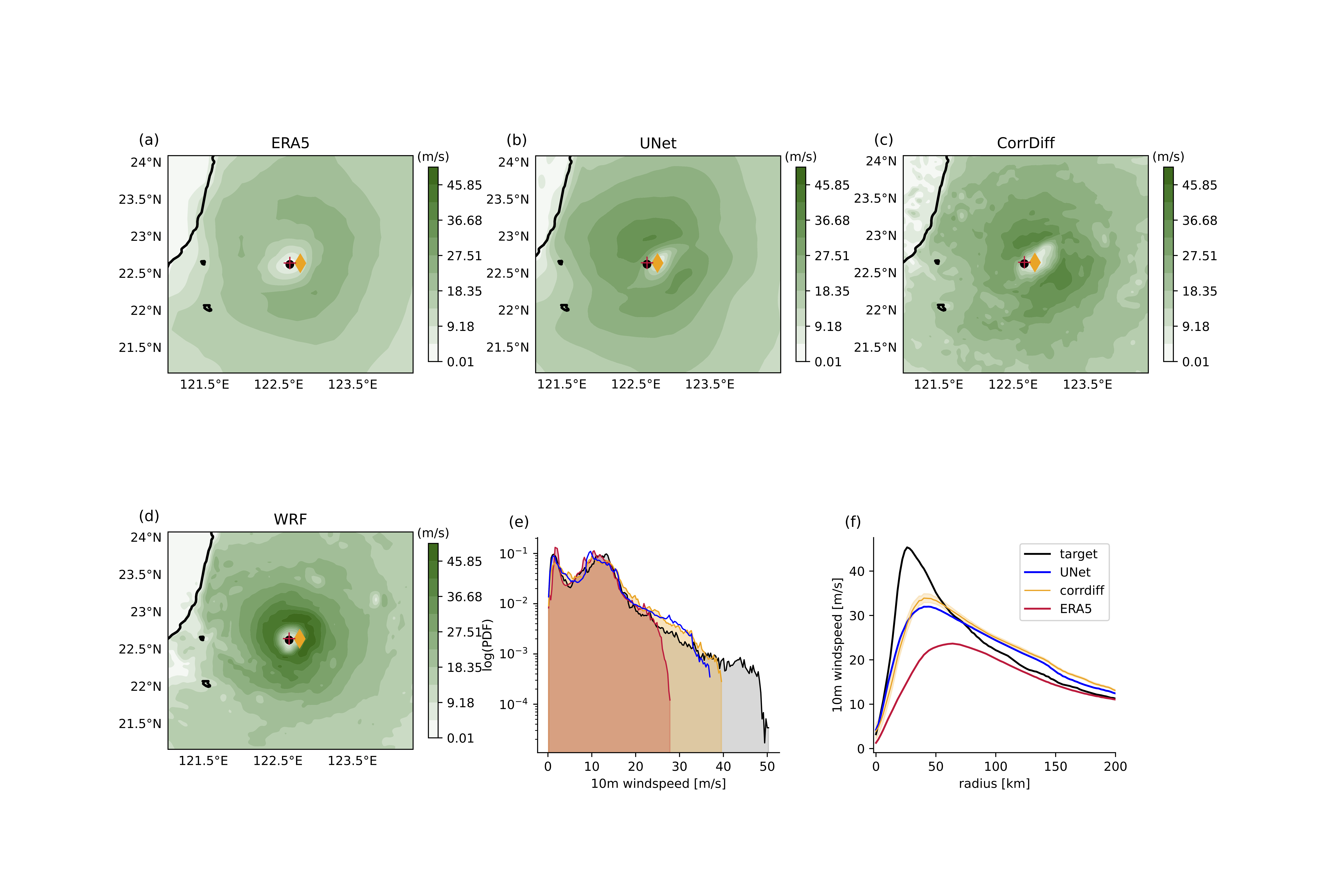}  
\caption{A comparison of the 10m windspeed maps ($m s^{-1}$), distributions and the axisymmetric cross section from typhoon Haikui (2023) on 2023-09-03 00:00:00 UTC. Panels (a),(b),(c), (d) show the 10m windspeed from ERA5, UNet, CorrDiff and target (WRF), respectively. The CorrDiff panels show the first of 32 ensemble members. The solid black contour indicates the Taiwan coastline. Storm center of the ERA5, CorrDiff and WRF are shown in red `+`, orange  diamond, and the black dot, respectively. Panels (e) shows the logarithm of the PDF of windspeed. Panel (f) shows the axisymmteric structure of the typhoon about its center, where for the CorrDiff curves, solid line shows the ensemble mean and the shading shows $\pm$ one standard deviation along the ensemble dimension.}  \label{fig:typhoon}
\end{figure}

An example illustrating the benefits and limitations of CorrDiff for downscaling typhoon Haikui (2023) is shown in Figure \ref{fig:typhoon}. Compared to the ground truth (Fig. \ref{fig:typhoon}d), the ERA5 reanalysis (Fig. \ref{fig:typhoon}a) poorly resolves the typhoon, depicting it as overly wide and with no closed contour annulus of winds above $16 m s^{-1}$. The UNet (Fig. \ref{fig:typhoon}b) likewise fails to recover a closed contour, although it does helpfully corrects approximately 50\% of error in the large-scale windspeed and structure compared to the target. CorrDiff (Fig. \ref{fig:typhoon}c) enhances the UNet by adding spatial variability, but maintains similar intensity.

The benefits of the CorrDiff downscaling compared to interpolating ERA5 can be more clearly quantified by examining the logarithm of the PDF of the windspeed, Fig.~\ref{fig:typhoon}(e). In the ERA5 the wind speed PDF has a sharp cutoff such that high wind speed values in excess of $27 m s^{-1}$ are missing. CorrDiff partially restores the tail of the typhoon wind speed PDF, and is capable of predicting wind speeds up to $ 40 m s^{-1}$ compared with the maximum value of $50 m s^{-1}$ in the target. The diffusion component of CorrDiff is responsible for the most extreme wind speeds in its predictions. In contrast, the mean axisymmetric structure of the typhoon (Fig.~\ref{fig:typhoon}f), is controlled more by the UNet, which reveals the influence of CorrDiff on typhoon geometry: With downscaling the radius of maximum winds decreases from 75km in ERA5 to about 50km in CorrDiff, compared with 25km in the WRF model. At the same time, the axisymmetric windspeed maximum increases from $22  m s^{-1}$ in ERA5 to $33 m s^{-1}$, compared with 45 in WRF -- both favorable improvements. Ultimately, CorrDiff is able to synthesis consistent radar reflectivity (see top row of Fig. \ref{fig:radar_ref}.  

For further discussion of typhoon downscaling, we refer the interested reader to  Appendix Section \ref{section:additional_typhoons}, where we explore additional date-times for Haikui (2023), investigate an additional typhoon Chanthu (2021), and analyze generated wind statistics across a 600-member ensemble of typhoon-containing time intervals spanning the 1980-2020 period. This extended analysis suggests that, while the main results emphasized for are case study above are qualitatively representative when typhoons are far from land, the diffusion component of CorrDiff frequently plays a stronger role in intensifying typhoon axisymmetric structure, and CorrDiff tends to lead to too much horizontal contraction of cyclone morphology, predicting a radius of maximum winds that is statistically too small.

\section{Discussion}
This study presents a generative diffusion model (CorrDiff) for multivariate downscaling of coarse-resolution (25-km) global weather states to higher resolution (2km) over a subset of the globe, and simultaneous radar channel synthesis. CorrDiff consists of two steps: regression and generation. The regression step approximates the mean, while the generation step further corrects the mean but also generates the distribution, producing fine-scale details stochastically. This approach is akin to the decomposition of physical variables into their mean and perturbations, common practice in fluid dynamics, e.g. \cite{pope2000turbulent}. 

Through extensive testing in the region of Taiwan, the model is shown to produce reasonably realistic power spectra and probability distributions of all target variables. The diffusion component of CorrDiff is found to be especially important for the task of radar channel synthesis.  Several case studies reveal that the model is able to downscale coherent structures consistently across its variables. Focusing on a midlatitude frontal event, horizontally co-located gradients of winds and temperatures are generated alongside spatially consistent radar reflectivity features, with incomplete but improved sharpness. For typhoons, encouraging partial corrections of typhoon size and wind speed intensity are found, alongside generated radar echos containing qualitatively realistic km-scale details reminiscent of tropical cyclone rainband morphology. It is logical to expect that the model's accuracy could be further improved with a larger training dataset that contains more diverse examples of such rare coherent structures such as by pre-training on large libraries of typhoons generated by high-resolution physical simulators; we encourage work in this direction.

Another important unsolved challenge is optimally calibrating CorrDiff's generated uncertainty to better match its error levels. This is somewhat unexpected since diffusion models for image generation are known to be over-dispersive in the sense of producing low-quality samples and variance-reducing techniques are often used during the sampling to discourage such outliers \cite{Ho2022-yb,Karras2024-ds}. The lack of grid-point-level spread here could owe to a number of factors in the diffusion training process including the noise schedules used, the comparatively large resolution (448x488) compared to typical image generation (64x64), or the weighting in the loss function.

To become useful for km-scale weather prediction, extensions of CorrDiff are encouraged that include temporal coherence, such as via video diffusion or learnt autoregressive km-scale dynamics; as with super-resolution these must be formulated as stochastic machine learning tasks. Currently, beyond the coherence of the large scale conditioning given from ERA5 there is no guarantee that CorrDiff's km-scale dynamics will be coherent in time. Additional integrations with km-scale data assimilation are also essential for this use case. Our current demonstration relies only on the global data assimilation (DA) used to produce the ERA5 dataset. Unlike the target data it is trained on, CorrDiff effectively bypasses the regional DA, which for CWA is of similar computational cost as running the operational numerical model (WRF) model for 13h. 

For such extensions, the two step approach in CorrDiff offers practical advantages to reduce the amount of variance that must be handled stochastically, and trade-off between the fast inference of the mean using the UNet, and the probabilistic inference of the CorrDiff. This is particularly useful given that some variables depend more than others on the diffusion step for their skill  (see Figure~\ref{fig:spectra_distributions}). Moreover, it could be possible to apply the diffusion step to a mean prediction obtained in a different way (e.g. a numerical model if available) to generate a plausible distribution from a single prediction. 

With the current hardware and code-base  CorrDiff inference is about $652$ times faster, and $1,310$ times more energy efficient than running CWA-WRF on CPUs, although such a comparison between dynamical and statistical downscaling is limited (see \ref{sec:speedup} for details). This paper focused on generation quality, and not on optimal inference speed, for which gains could be easily anticipated. Our CorrDiff prototype is using a dozen iterations thanks to the initial regression step. Refinement of the technique could reduce the number of iterations to only a few by using distillation methods~\cite{salimans2022progressive,xiao2022DDGAN, zheng2023fast} and pursuing other performance optimization techniques~\cite{meng2023distillation,vahdat2021score}.

If some of the above challenges in the model are resolved, several potential extensions of the proposed method are worth consideration by the community:

\begin{enumerate}
    \item \textbf{Downscaling Coarse-Resolution Medium-Range Forecasts:} 
    This requires addressing lead time-dependent forecast errors in the input, enabling a comprehensive evaluation of simultaneous bias correction and downscaling, and adding temporal coherence and km-scale prediction and data assimilation capabilities.

    \item \textbf{Downscaling at Different Geographic Locations:} The primary obstacle here is the scarcity of reliable kilometer-scale weather data. Additionally, addressing the computational scalability of CorrDiff for regions significantly larger than Taiwan is crucial.

    \item \textbf{Downscaling Future Climate Predictions:} This introduces further complexities related to conditioning probabilistic predictions on various future anthropogenic emissions scenarios and assessing whether the generated weather envelope appropriately reflects climate sensitivity, particularly concerning extreme events.

    \item \textbf{Synthesizing sub-km sensor observations:} To achieve effective resolutions beyond what is possible to simulate today, and sidestep issues of numerical simulation, it would be interesting to test whether variants of CorrDiff can be trained to generate raw senor observations where dense networks exists. Our demonstrated ability to synthesize an observable as challenging as radar reflectivity from column water vapor should embolden such efforts. 

\end{enumerate}

These extensions have significant potential benefits such as accelerated regional forecasts, increased ensemble sizes, improved climate downscaling, and the provision of high-resolution regional forecasts in data-scarce regions, leveraging training data from adjacent areas.

\section{Methods}

\label{sec:methods}
This section elaborates on the proposed CorrDiff methodology for probabilistic downscaling. It begins with a background on diffusion models to provide the machinery. It then delves into CorrDiff and its associated components. We further detail our experimental setup including the CWA dataset, network architecture, and training protocols. At the end, we briefly discuss evaluation criteria.

\subsection{Background on diffusion models}
Consider the data distribution represented by \(p_{{\rm data}}({\bf x})\). This distribution has an associated standard deviation, denoted by \(\sigma_{{\rm data}}\). The forward diffusion process seeks to adjust this distribution, yielding modified distributions denoted by \(p_{{\rm data}}({\bf x}; \sigma)\). This transformation is achieved by incorporating i.i.d. Gaussian noise with a standard deviation of \(\sigma\) into the data. When \(\sigma\) surpasses \(\sigma_{{\rm data}}\) by a considerable margin, the resulting distribution approximates pure Gaussian noise.

Conversely, the backward diffusion process operates by initially sampling noise, represented as \({\bf x}_0\), from the distribution \(\mathcal{N}(0, \sigma_{\max}^2 \bf{I})\). The process then focuses on the denoising of this image into a series, \({\bf x}_i\), that is characterized by a descending order of noise levels: \(\sigma_0=\sigma_{\max} > \sigma_1 > \ldots > \sigma_N=0\). Each noise level corresponds to a specific distribution of the form \({\bf x}_i \sim p_{{\rm data}}({\bf x}_i;\sigma_i)\). The terminal image of the backward process, \({\bf x}_N\), is expected to approach the original data distribution.

\noindent\textbf{formulation of the underlying stochastic differential equations}. To present the forward and backward processes rigorously, they can be captured via stochastic differential equations (SDEs). Such SDEs ensure that the sample, \(\bf x\), aligns with the designated data distribution, \(p\), over its progression through time \cite{song2021scorebased, karras2022elucidating}. A numerical SDE solver can be used here, where a critical component is the noise schedule, \(\sigma(t)\), which prescribes the noise level at a specific diffusion-time, \(t\). Here 'diffusion time' is a virtual variable used for denoting denoising steps and has roots in differential equations that are derived from Langevin dynamics. To avoid confusion with the time of day, we use diffusion time to denote this variable. A typical noise scheduler is \(\sigma(t) \propto \sqrt{t}\). Based on \cite{karras2022elucidating}, the forward SDE is given as
%
\begin{align}
    d{\bf x} =  \sqrt{2 \dot{\sigma}(t) \sigma(t)} d \pmb{\omega}(t),   \label{eq:forward_sde}
\end{align}
while the backward SDE is
%
\begin{align}
    d{\bf x} = - 2 \dot{\sigma}(t) \sigma(t) \nabla_{{\bf x}} \log p({\bf x};\sigma(t)) dt + \sqrt{2 \dot{\sigma}(t) \sigma(t)} d \bar{\pmb{\omega}}(t).    \label{eq:backward_sde}
\end{align}
The term \(\dot{\sigma}(t)\) refers to the derivative of \(\sigma(t)\) with respect to the diffusion-time. Here $\omega$ in the forward SDE is a Wiener process, while the backward SDE comprises two terms: a deterministic component representing the probability flow ODE with noise degradation, and noise injection via the Wiener process. 

\noindent\textbf{Denoising score matching}. An examination of the SDE in Eq.~\eqref{eq:backward_sde} indicates the necessity of the score function, \(\nabla_{{\bf x}} \log p({\bf x};\sigma)\), for sampling from diffusion models. Intriguingly, this score function remains unaffected by the normalization constant of the base distribution, regardless of its computational complexity. Given its independence, it can be deduced using a denoising method. If \(\nabla_{{\bf x}} \log p({\bf x};\sigma) = (D_{\theta}({\bf x}; \sigma) - {\bf x})/\sigma^2\), a denoising neural network, namely $D_{\theta}({\bf x}; \sigma)$, can be trained for the denoising task using
\begin{align}
    \min_{\theta} \mathbbm{E}_{{\bf x} \sim p_{\rm data}}   \mathbbm{E}_{{\sigma} \sim p_{\sigma}}\mathbbm{E}_{{\bf n} \sim \mathcal{N}(0, \sigma^2 {\bf I})} \big[ \|D_{\theta}( {\bf x}+ {\bf n}; \sigma) - {\bf x}\|^2  \big].   \label{eq:score_matching}
\end{align}
Note, the noise variance is also modeled as a random variable that simulates different noise levels in the forward process e.g., based on log-normal distribution; see \cite{karras2022elucidating}.

\subsection{Proposed approach}
As discussed in section \ref{sec:generativedownscaling}, the Fig.  state ${\bf{x}}$ in \eqref{eq:decomposition} can be written as the sum of the mean $\boldsymbol{\mu}$ and the difference $\bf{r}$, where the latter will be nearly zero mean and exhibits a small distribution shift, which facilities training diffusion models for correcting the mean prediction. It is worth noting that this two-step method has further implications for learning physics. The UNet-regression step can anticipate many of the physics of downscaling, some of which are deterministic. These include high-resolution topography (which to first order controls the 2-meter temperature variation due to the lapse-rate effect), and the large-scale horizontal wind which combine leading balances in the free atmosphere with the effect of surface friction and topography. Stochastic phenomena such as convective storms that also change temperatures and winds are easier to model as deviations from the mean. Also, cloud resolving models are explicitly formulated using deviations from a larger scale balanced state \cite{pressel2015large}. In the next section, we discuss the regression and generation step in details.

\subsubsection{Regression on the mean}
In order to predict the conditional mean ${\pmb{\mu}} = \mathbbm{E} [{\bf x}|{\bf y}]$, we employ to a UNet-regression model trained on a dataset $\{({\bf x}_n, {\bf y}_n)\}_{n=1}^N$. We utilize the specific architecture described in ~\cite{song2019generative} that incorporates attention and residual layers, allowing it to effectively capture both short and long-range dependencies in the data (see section \ref{sec:architecture-description} and \ref{fig:UNet}).  The model is optimized using  a Mean-Squared-Error (MSE) loss during training.

\subsubsection{Denoising diffusion corrector}
\label{sec:Denoising}
Once equipped with the UNet-regression network, we can begin by predicting the conditional mean $\hat{\pmb\mu}$, which serves as an approximation of  $\mathbbm{E} [{\bf x}|{\bf y}]$. Subsequently, we proceed to train the diffusion model directly on the difference: ${\bf r} = {\bf x} - \hat{{\pmb\mu}}$. Notably, the difference exhibits a small departure from the target data, allowing for the utilization of smaller noise levels during the training of the diffusion process.

In our approach, we adopt the Elucidated diffusion model (EDM), a continuous-time diffusion model that adheres to the principles of SDEs (in Eq.~\eqref{eq:forward_sde}-\eqref{eq:backward_sde}) \cite{karras2022elucidating} to design the diffusion process and architecture. As a result it has an intuitive and physics driven hyperparameter tuning, which makes it work across different domains. In our case, we want to generate ${\bf r}$ by sampling from the conditional distribution $p({\bf r} | {\bf y})$ following the SDEs in Eq.~\eqref{eq:forward_sde}-\eqref{eq:backward_sde}. To condition the diffusion model, we concatenate the input coarse-resolution data ${\bf y}$ with the noise over different channels. We also learn the score function $\nabla_{{\bf r}} \log p({\bf r} | {\bf y})$ using the score matching loss in Eq.~\eqref{eq:score_matching} where the denoiser is now $D_{\theta}({\bf r}+{\bf n}; \sigma; {\bf y})$ with the conditioning input ${\bf y}$. For the denoiser we again follow the design principles in EDM to use a UNet architecture with skip connections weighted by the noise variance. Architecture details are discussed in Section~\ref{sec:architecturedetail}.

To generate samples from the distribution $p({\bf r}|{\bf y})$, we employ the second-order EDM stochastic sampler \cite{karras2022elucidating} [Algorithm 2] to solve for the reverse SDE in Eq.~\eqref{eq:backward_sde}. Upon sampling ${\bf r}$, we augment it with the predicted conditional mean $\hat{\pmb\mu}$ from regression, to generate the sample $\hat{\pmb\mu} + {\bf r}$. This entire workflow is illustrated in Fig.~\ref{fig:CorrDiff_diagram}, providing a visual representation of the steps involved in our proposed method.

\subsection{Experimental setup}

\subsubsection{Dataset}
Our training and test data cover the region of Taiwan and surrounding ocean. The choice of region is driven by our partnership with a local government agency (CWA) which operates a regional dynamical downscaling. However, the region of Taiwan also presents a unique diversity of meteorological conditions and phenomena such as tropical cyclones (i.e., typhoons), mid-latitude cyclones which generate weather fronts, and steep topography with snow-capped mountains and land-sea contrast. Such diversity is hard to find elsewhere in such a relatively small domain.

The input (conditioning) dataset is taken from  ERA5 reanalysis \cite{hersbach2020era5}, a global dataset at spatial resolution of about 25km and a temporal resolution of 1h, the latter matches the target data listed below. To facilitate training, we interpolate the input data onto the curvilinear grid of CWA with bilinear interpolation (with a rate of 4x), which results in $36 \times 36$ pixels over the region of Taiwan. Each sample in the input dataset consists of 12 channels of information; see Table \ref{tab:input_output}. This includes two pressure levels (500 hPa and 850 hPa) with four corresponding variables: temperature, eastward and northward components of the horizontal wind vector, and geopotential height. Additionally, the dataset includes single-level variables such as 2-meter Temperature, 10-meter wind vector and total column water vapor. This input channel set is admittedly somewhat arbitrary, but serves the purposes of (i) allowing a reasonable sparse representation of the thermodynamic state of the macro-scale atmosphere, while (ii) intentionally including only limited information about atmospheric water via the total vertical integral of an invisible trace gas - water vapor. This ensures the task of synthesizing radar reflectivity - a much more complex observable that relates to the sixth moment of the cloud water droplet distribution - is appropriately ambitious, as a strong test of CorrDiff's generative component.

The target dataset used in this study is a subset of the proprietary RWRF model data (Radar Data Assimilation with WRFDA \footnote{https://www.mmm.ucar.edu/models/wrfda}). The RWRF model is one of the operational regional Numerical Weather Prediction (NWP) models developed by CWA \cite{chen2020improving}, which focuses on radar Data Assimilation (DA) in the vicinity of Taiwan. Assimilating radar data is a common strategy used in regional weather prediction, which helps constrain especially stochastic convective processes such as mesoscale convective systems and short-lived thunderstorms. In addition, CWA assimilates several surface measurements to complement the radar data that often miss the surface values. The WRF-CWA system uses a nested 2km domain in a larger 10km domain that is driven by a global model (GFS) as boundary conditions \cite{chen2020improving}. By incorporating radar data \cite{chang2021operational}, RWRF improves the short-term prediction of high-impact weather events. The radar observations possess high spatial resolution of approximately 1km and temporal resolutions of 5-10 minutes at a convective scale. These observations provide useful wind information (radial velocity) as well as hydrometeors (radar reflectivity), with a particular emphasis on the lower atmosphere. The radar data assimilation relies on the availability of reliable and precise observations, which contributes significantly to enhance the accuracy and performance of the applied deep learning algorithms in the context of NWP applications.

The target dataset covers the years 2018-2021. It has a temporal frequency of one hour and a spatial resolution of 2km. We use only the inner (nested) 2km domain, which has $448 \times 448$ pixels, projected using the Lambert conformal conical projection around Taiwan. The geographical extent of the dataset spans from approximately 116.371°E to 125.568°E in longitude and 19.5483°N to 27.8446°N in latitude. We sub-selected 4 channels (variables) as the target variables that are most relevant for practical forecasting: temperature at 2 meter above the surface, the horizontal winds at 10 meter above the surface and the 1h maximum derived radar reflectivity (radar reflectivity hereafter) - a surrogate of the expected precipitation. Notably, the radar reflectivity channel is not present in the input data, and needs to be predicted based on the other channels (channel synthesis). The radar reflectivity data also exhibits a distinct distribution compared to the other output channels, with positively skewed values and a prominent zero-mode consistent with typical non-raining conditions. 

Initially, the target data is provided in the NetCDF format, which is the output of the WRFDA assimilation process. A vertical interpolation from hybrid coordinates (i.e., sigma levels) to pressure coordinates (i.e., isobaric levels) is applied. As part of the preprocessing steps, the data is converted to the Hadoop Distributed File System (HDFS) format. 
Additionally, any missing or corrupted data points represented by "inf" or "nan" values are eliminated from the dataset. This leads to a reduction in the number of images from 37,944 to 33,813. See \ref{sec:architecture-description} for more details.

To avoid over-fitting, we divide the data into training and testing sets. Three years of data 2018-2020 are used for training (24,154 images total) and 2021 is used for testing. Some selected dates from 2022 and 2023 are used for case studies.

\subsubsection{Network architecture and training}
\label{sec:architecturedetail}
The CorrDiff method has two step learning approach. To ensure compatibility and consistency, we employ the same UNet architecture for both the regression and diffusion networks. We enhance the UNet of \cite{song2019generative} by increasing its size to include 6 encoder layers and 6 decoder layers. The base embedding size is set to 128, and it is multiplied over channels according to the multipliers [1,2,2,2,2]. The attention resolution is defined as 28. To represent \textit{time} (i.e., timesteps in the diffusion process, not to be confused with the time of day), we utilize the Fourier-based position embedding. However, in the regression network, time embedding is disabled as no probability flow ODE is involved. No data augmentation techniques are employed during training. Overall, the UNet architecture comprises $80$ million parameters. Additionally, we introduce $4$ channels for sinusoidal positional embedding to improve spatial consistency, following established practices in the field \cite{vaswani2017attention, dosovitskiy2020image, carion2020end}.

During the training phase, we use the Adam optimizer with a learning rate of $2 \times 10^{-4}$, $\beta_1=0.9$, and $\beta_2=0.99$. Exponential moving averages (EMA) with a rate of $\eta=0.5$ are applied, and dropout with a rate of 0.13 is utilized. We adopt the hyperparameters based on the guidelines proposed in EDM, Karras et al., (2022), including the same optimizer hyperparameters and learning rate schedule. EDM offers a physics-inspired design space based on ODEs that can be auto-tuned to our scenario (see table 1 in \cite{karras2022elucidating}). The only relevant hyperparameter that is tenable in this framework is $P_{mean}$ of the noise schedule. The value of this parameter was selected based on the resolution and dynamic range of the data as done in \cite{hoogeboom2023simple}.

The regression network receives only the $12$ input channels from the ERA5 conditioning data. In contrast, the diffusion training concatenates these same $12$ input conditioning channels from the coarse-resolution ERA5 data with 4 noise channels to generate the output for each denoiser. To further enhance the diffusion conditioning, we also add the mean obtained by the regression model in the first stage. This addition provides more context and global information to the diffusion process, potentially improving its convergence.. For diffusion training, we adopt the Elucidated Diffusion Model (EDM), a continuous-time diffusion model. During training, EDM randomly selects the noise variance such that $\ln(\sigma(t)) \sim \mathcal{N}(0,1.2^2)$ and aims to denoise the samples per mini-batch. EDM is trained for 28 million steps, while the regression UNet is trained for only 2 million steps. The training process is distributed across 16 DGX nodes, each equipped with 8 H100 GPUs, utilizing data parallelism and a total batch size of 512. The total training time for regression and diffusion models was 7 days that amounts to approximately $21,504$ GPU-hours.

For the residual diffusion process, during training we adopt log-normal distribution for the noise standard-deviation $\sigma$; see \eqref{eq:score_matching} \cite{karras2022elucidating}. We choose $\sigma \sim {\rm lognormal} (\mu=0.0, \sigma=1.2)$ to ensure the forward diffusion completely destructs the large data intensity especially for the radar reflectivity variable. For sampling purposes, we employ the second-order stochastic sampler provided by EDM. This sampler performs $18$ steps, starting from a maximum noise variance of $\sigma_{\max}=800$ and gradually decreasing it to a minimum noise variance of $\sigma_{\min}=0.002$. We adopt the rest of hyperparamaters from EDM as listed in \cite{karras2022elucidating}.

\subsection{Evaluation criterion}
Probabilistic predictions aim to maximize sharpness subject to calibration \cite{Raftery2005-md}.
Qualitatively, calibration means that the likelihood of observing the true value is the same as observing a member drawn from the ensemble. A necessary condition for calibration is that the spread-error relationship be 1-to-1 when averaged over sufficient samples \cite{Fortin2014-fc}. Calibration also manifests as a flat rank-histogram, both are reported in \ref{sec:calibration}. A simple metric used below is the root-mean-squared error of the sample mean. In the large sample limit, the sample mean becomes deterministic. So we expect this error to be comparable for generative and deterministic models.

Instead of considering both calibration and spread separately, it can be easier to use proper scoring rules like the continuous-ranked-probability score (CRPS) \cite{Gneiting2007-rq}.
Let $x$ be a scalar observation and $F$ be the cumulative distribution of the probabilistic forecast (e.g., the empirical CDF of generated samples). Then, CRPS is defined as 
\[
    CRPS(F, x) = \int_{-\infty}^{\infty} (F(y) - \mathbbm{1}_{\{y \geq x\}})^2 \, dy,
\]
here  $\mathbbm{1}_{\{y \geq x \}}$ is the Heaviside step function, and $F$ which minimizes CRPS is the true cumulative distribution of $x$.
For a deterministic forecast, $F(y)=\mathbbm{1}_{\{y \geq x_0}\}$  where $x_0$ is the forecast value, CRPS is equivalent to the mean absolute .

\section{Acknowledgements}
We extend our profound appreciation to the Central Weather Administration (CWA) of Taiwan\footnote{https://www.cwa.gov.tw/eng/}, a premier government meteorological research and forecasting institution, for granting us access to the invaluable operational Numerical Weather Prediction (NWP) model dataset and for their expert guidance on data consultation. Our gratitude also extends to the AI-Algo team at NVIDIA, especially Kamyar Azizzadenesheli, Anima Anandkumar, Nikola Kovachki, Jean Kossaifi, and Boris Bonev for their insightful discussions. Additionally, we are indebted to David Matthew Hall, Dale Durran, Chris Bretherton for their constructive feedback on the manuscript.

\bibliographystyle{plain}
\bibliography{refs}

\clearpage
\beginsupplement
\setcounter{section}{0}
\setcounter{subsection}{0} 
\section*{Supplementary Information}

\section{Our position with respect to existing works}
\label{related_works}
To highlight the novel component of our work, we provide an expanded review of relevant literature. Table \ref{tab:related_works} presents a shortlist of the most relevant works that perform weather downscaling. 
Previous ML downscaling efforts have achieved notable successes in various areas. These include state vector inflation as \cite{addison2022machine}, application to large spatial domains \cite{harris2022generative}, achieving a large resolution ratio, e.g. \cite{hatanaka2023diffusion, addison2022machine} and downscaling of precipitation in tropical cyclones \cite{vosper2023deep}.
It is worth noting, however, that the majority of these studies concentrate on downscaling a single variable per model (note that \cite{leinonen2020stochastic} provided two models, each for a single variable, and is thus listed twice in the table). The variables of interest in all these works primarily relate to cloud and precipitation properties. \cite{vosper2023deep} showed a successful super resolution (recovering 10km from data coarsened to 100km) of tropical cyclone precipitation. Despite these advancements,  to the best of our knowledge ML downscaling that accounts for different physics, across many channels and channel synthesis in a single model was not shown before. 

The combined prediction of selected dynamical, thermodynamical and microphysical (cloud related) variables in concert marks a new capability of such models. Its utility is demonstrated here by examining coherent structures and how all variables jointly downscaled in a physically consistent manner. 

\begin{table}
\renewcommand{\arraystretch}{1.5} 
\centering
\begin{tabular}{|p{4.1cm}|p{1.8cm}|p{2.8cm}|p{1.0cm}|p{4.3cm}|} 
\toprule[0.15em] 
Citation & Architecture & Resolutions & Pixels & Variables \\
\hline 
Addison et al., (2022) \cite{addison2022machine} & Diffusion & input: 60km \newline target: 8.8km & $64^2$ & precipitation \\
\hline
Harris et al., (2022) \cite{harris2022generative} & GANs + \newline  ensemble  \newline  forecast & input: 10km \newline  target: 1km & $940^2$ & precipitation \\
\hline
Hatanaka et al., (2023) \cite{hatanaka2023diffusion} & Cascaded \newline  diffusion & input: 30km \newline target: 1km & $128^2$ & day-ahead  \newline solar-irradiance \\
\hline
Leinonen et al., (2020) \cite{leinonen2020stochastic} & GANs &  input 8km \newline target: 1km & $128^2$ & precipitation\\
\hline
Leinonen et al., (2020) \cite{leinonen2020stochastic} & GANs &  input 16km \newline target: 2km & $128^2$ & cloud optical-thickness \\
\hline
Price and Rasp, (2022) \cite{price2022increasing} & Corrector GAN & input 32km \newline target 4km & $128^2$ & precipitation \\
\hline
Vosper et al., (2023) \cite{vosper2023deep} & WGAN & input 100km \newline target 10km & $100^2$ & precipitation from \newline tropical-cyclones\\
\hline
Current work  & CorrDiff & input 25km \newline target 2km & $448^2$ & 10-meter winds \newline 2-meter temperature \newline radar-reflectivity\\

\bottomrule[0.15em] 
\end{tabular}
\caption{Downscaling models presented in the most relevant works we could find with respect to the current study. We  highlight the resolution ratios, the pixel size of the Fig.  prediction, predicted variables and architecture.}
\label{tab:related_works}
\end{table}

\section{Descriptions of the architecture and the training data}
\label{sec:architecture-description}

\begin{figure}
    \centering
    \includegraphics[width=1.1\textwidth,clip]{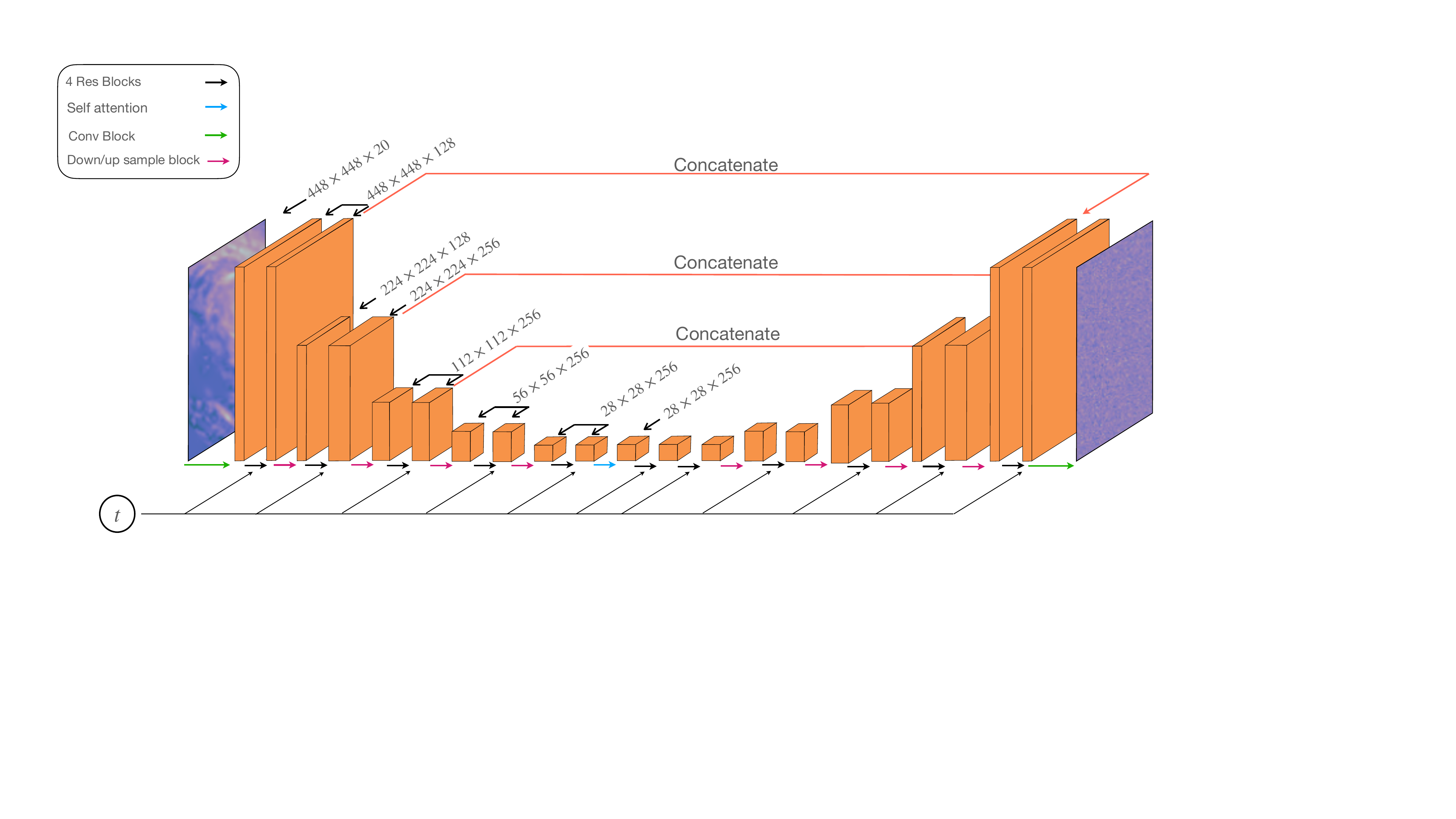}
    \caption{A sketch of the hierarchical UNet architecture adopted in both the regression model and the denoising diffusion model. Note, in the regression stage, time embedding is not used. }
    \label{fig:UNet}
\end{figure}

Figure \ref{fig:UNet} illustrates the architecture of the UNet, which serves a dual purpose in CorrDiff: it functions as both the regression model and the denoiser in the diffusion model. Table \ref{tab:input_output} provides a comprehensive list of the input and output channels utilized by CorrDiff. It is important to note that these input and output datasets differ not only in their pixel size but also in the specific channels they encompass. For single-level variables, the input includes total column water vapor but lacks the maximum radar reflectivity, which is present in the output, and vice versa. The input also includes pressure level variables at the $850$ and $500$ (hPa) levels, which combine to 12 input channels.

\begin{table}[ht]
\centering
\begin{tabular}{|l|l|l|}
\hline
 & \textbf{Input} & \textbf{Output} \\
\hline
\textbf{Pixel side} & 36 x 36 & 448 x 448 \\
\hline
\textbf{Single level channels} & Total column water vapor & Maximum radar reflectivity \\
 & Temperature at 2 meter & Temperature at 2 meter  \\
 & Eastward wind at 10 meter & Eastward wind at 10 meter \\
 & Northward wind at 10 meter & Northward wind at 10 meter  \\
\hline
\textbf{Pressure level channels} & Temperature &  \\
 & Geopotential &  \\
 & Eastward wind &  \\
 & Northward wind &  \\
\hline
\end{tabular}
\caption{A list of the input and output resolutions and channel for the CorrDiff downscaling model. Input channels include the both single level variables and pressure level variables, the latter are used at  $850$ and $500$ (hPa) levels.}
\label{tab:input_output}
\end{table}

Figure \ref{fig:stats} below shows a time series of the normalized target data, where the mean and the 2 and 4 standard deviations from the mean are plotted. A single corrupted data point is evident by the negative spike in 2m temperature. Some missing data periods are also evident by gaps, such as around May 2021; these were removed due to 'NaNs' or 'Inf' in either input or target data. The validation data constitutes the last 12 months from 2021-01-01 00:00:00 UTC. Moreover, the training data (2018-2020) has 22 hourly snapshots from 5 named typhoons of category 1 or higher in the domain. As a rough estimate, frontal systems seem to pass through the domain between 4-8 time per year (every 2-3 weeks in the winter) and typically last about 8h within the domain. Therefore, both typhoons and frontal systems are rather rare in the training data. \\

\begin{figure}
    \centering
        \hspace{0mm}\includegraphics[width=0.6\textwidth]{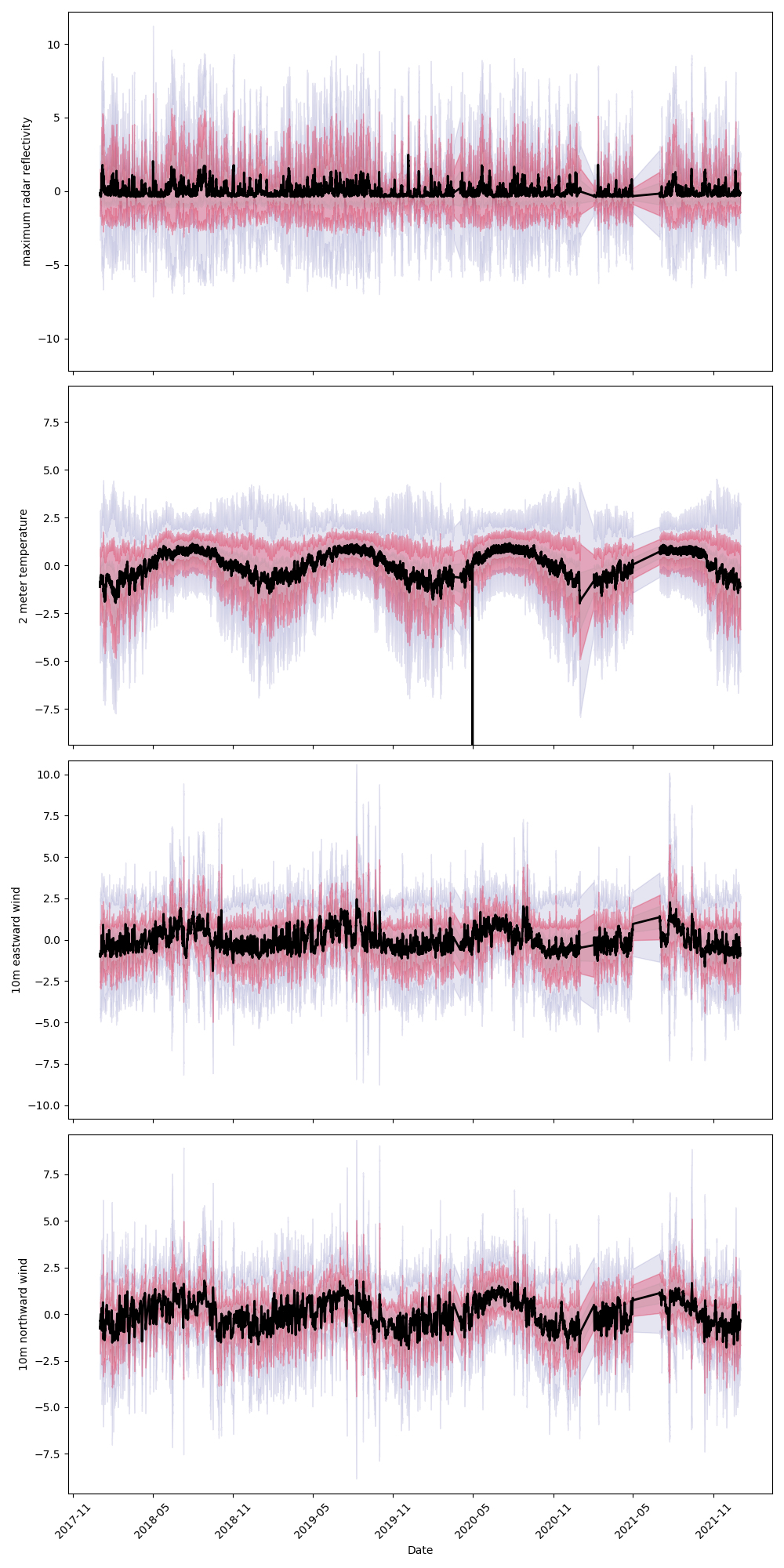}  
\caption{Time series of the mean (black line), mean $\pm 2$ standard deviation (red shading) and mean $\pm 4$ standard deviation (grey shading) for the four target channels as used by the model. This analysis is after re-scaling the channels by subtracting their global mean and dividing by global standard deviation. Statistics is computed over the domain at each date and time in the dataset.}  \label{fig:stats}
\end{figure}

\section{Localization by two-step formulation}
\begin{figure}
    \centering
    \includegraphics[width=0.8\textwidth,clip]{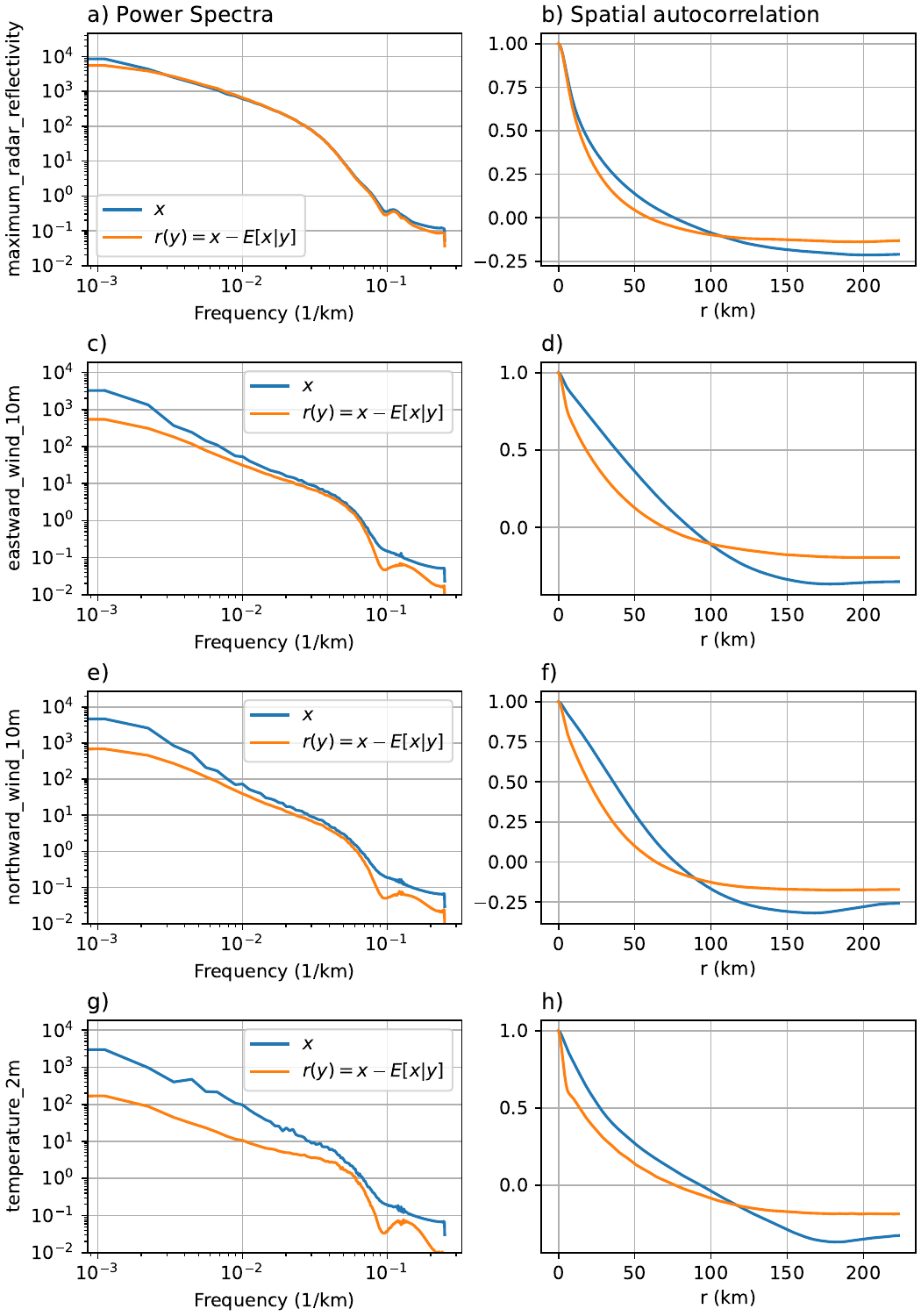}
    \caption{Left column: power spectra, right column:  spatial auto-correlation. Top to bottom: maximum radar reflectivity, 10m eastward wind, 10m northward wind and 2m temperature. This figure compares the original target $x$ and the difference $r=x-\mathbbm{E}[x|y]$. The difference has reduced variance at large-scales and equivalently removes the long-range auto-correlations.}
    \label{fig:acf}
\end{figure}

Figure \ref{fig:acf} (left) demonstrates the role of the two steps associated with CorrDiff as a function of spatial scales. From the target data (blue), it is seen that the regression step learns larger spatial scales, leaving some of the small scales for the diffusion step. In addition, from Fig. \ref{fig:acf} (right), it is observed that the residual is more localized and varies less overall, especially for the temperature field, which is strongly driven by changes in elevation. This has important implications for training and sampling efficiency of diffusion models as one can deploy diffusion models, with smaller UNet denoising architectures to aggregate the local information. We leave further study of this for future research.

\label{sec:acf}

\section{Examining sample diversity of CorrDiff}
\label{sample_diversity}
In order to showcase the realism of individual samples from CorrDiff and their quality compared with the target data, Fig. \ref{fig:animation} shows an animation of the target data (left), UNet (middle), and 20 generated samples of the CorrDiff prediction (right) of maximum radar reflectivity.

\begin{figure}[htbp]
  \centering
  \animategraphics[width=\linewidth,autoplay,loop]{10}{animation/000}{000}{159}
  \caption{Comparative analysis of maximum radar reflectivity across multiple samples: Diffusion model predictions versus UNet regression and WRF target simulations for diverse cloud regimes. This figure is presented as an animated visualization, viewable as a video in Adobe-compatible formats. The full animation can be accessed and downloaded from \url{https://figshare.com/ndownloader/files/48060031}.}
  \label{fig:animation}
\end{figure}

\section{Pooled metrics}
\label{sec:pooled_metrics}
Due to the stochastic nature of km atmospheric fields, and specifically radar reflectivity, we complement the main text with a pooled counterpart of the metrics reported in \ref{tab:big-score}. In \ref{tab:pooled_scores} the MAE and CRPS are computed from data data pooled over 14 points, which is roughly the resolution of the ERA5 data used for conditioning. 

\begin{table}
\centering
\begin{tabular}{@{}llrrrr@{}}
\begin{tabular}{lrrrr}
\toprule
 & Radar & t2m & u10m & v10m \\
\midrule
CorrDiff (CPRS) & 1.64 & 0.5 & 0.73 & 0.82 \\
CorrDiff (MAE) & 2.0 & 0.59 & 0.88 & 0.98 \\
UNet (MAE) & 1.98 & 0.53 & 0.89 & 0.99 \\
RF (MAE) & 2.91 & 0.62 & 0.91 & 1.01 \\
ERA5 (MAE) & - & 0.83 & 0.97 & 1.05 \\
\bottomrule
\end{tabular}
\end{tabular}
\vspace{2mm}
\caption{pooled CRPS counterpart table for \ref{tab:big-score}, where the prediction and target from each model are pooled on 28 grid boxes.}
\label{tab:pooled_scores}
\end{table}

Here as well, CorrDiff CRPS is superior to the MAE for all baselines, followed by the UNet, RF and ERA5 interpolation. This implies that the results presented in the main text are a valid reflection of the model's performance compared with the baselines.

\section{Additional Case Study Analysis}
\label{section:additional_cases}

To complement the example we have in the main text, we provide here an additional analysis of the case studies. For weather fronts, we analyze the collocation of the front with the reflectivity (clouds and rain), and for typhoons, we examine two different typhoons as well as historical typhoons for which there is no target data.

\subsection{Additional Weather Front Analysis}
\label{section:additional_front}
\begin{figure}
    \centering        \hspace{0mm}\includegraphics[width=1\textwidth,clip]{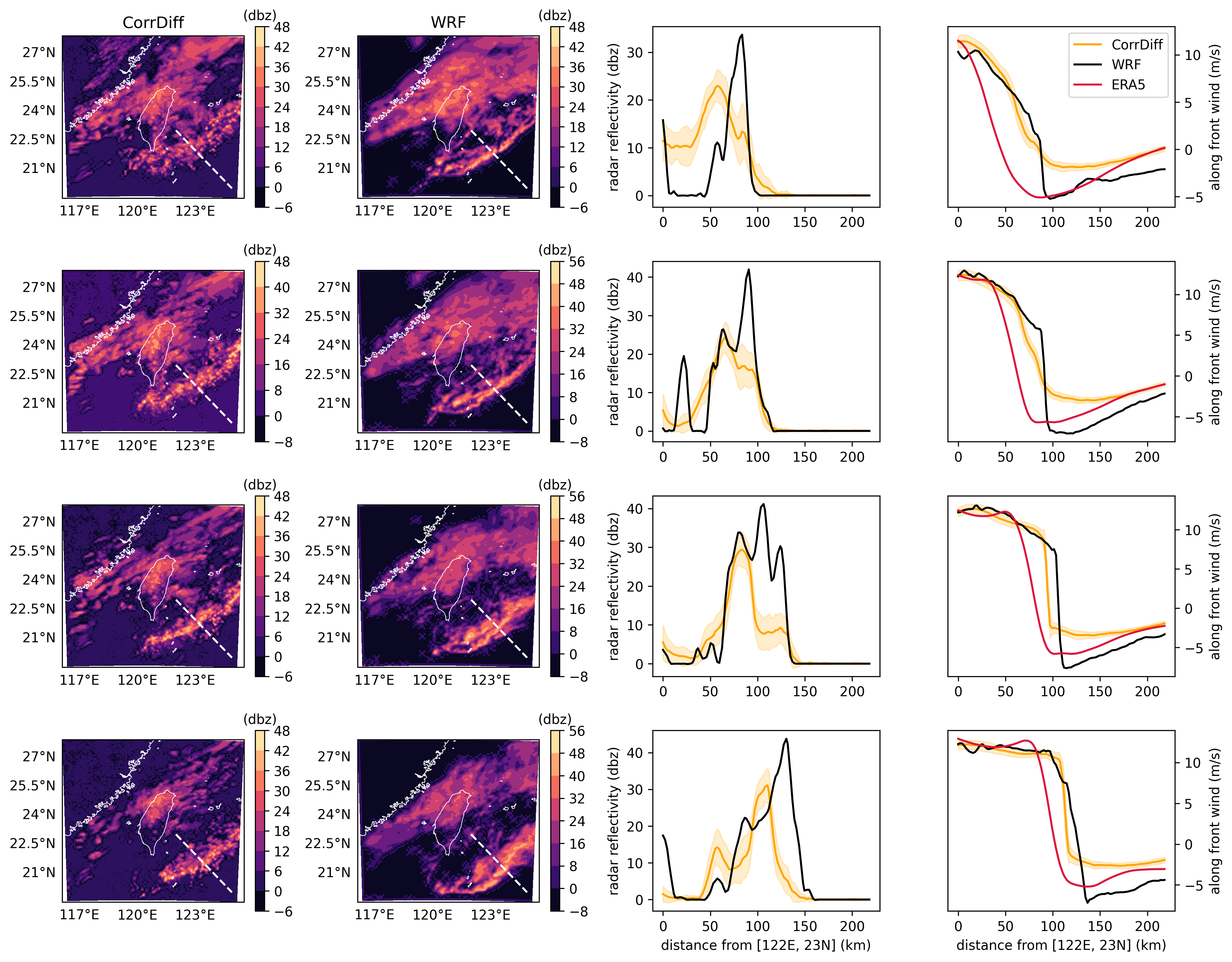} 
\caption{Analyzing the reflectivity synthesized during a cold front. Left to right: radar reflectivity maps of CorrDiff and the target data, followed by the transects of reflectivity and of the along front wind. along the dashed line in the adjacent maps. Top to bottom:2022-02-13 17:000:00, 2022-02-13 19:00:00, 2022-02-13 21:00:00 and 2022-02-13 23:00:00 UTC.}  \label{fig:front_ref_2022}
\end{figure}

\begin{figure}
    \centering        \hspace{0mm}\includegraphics[width=1\textwidth,clip]{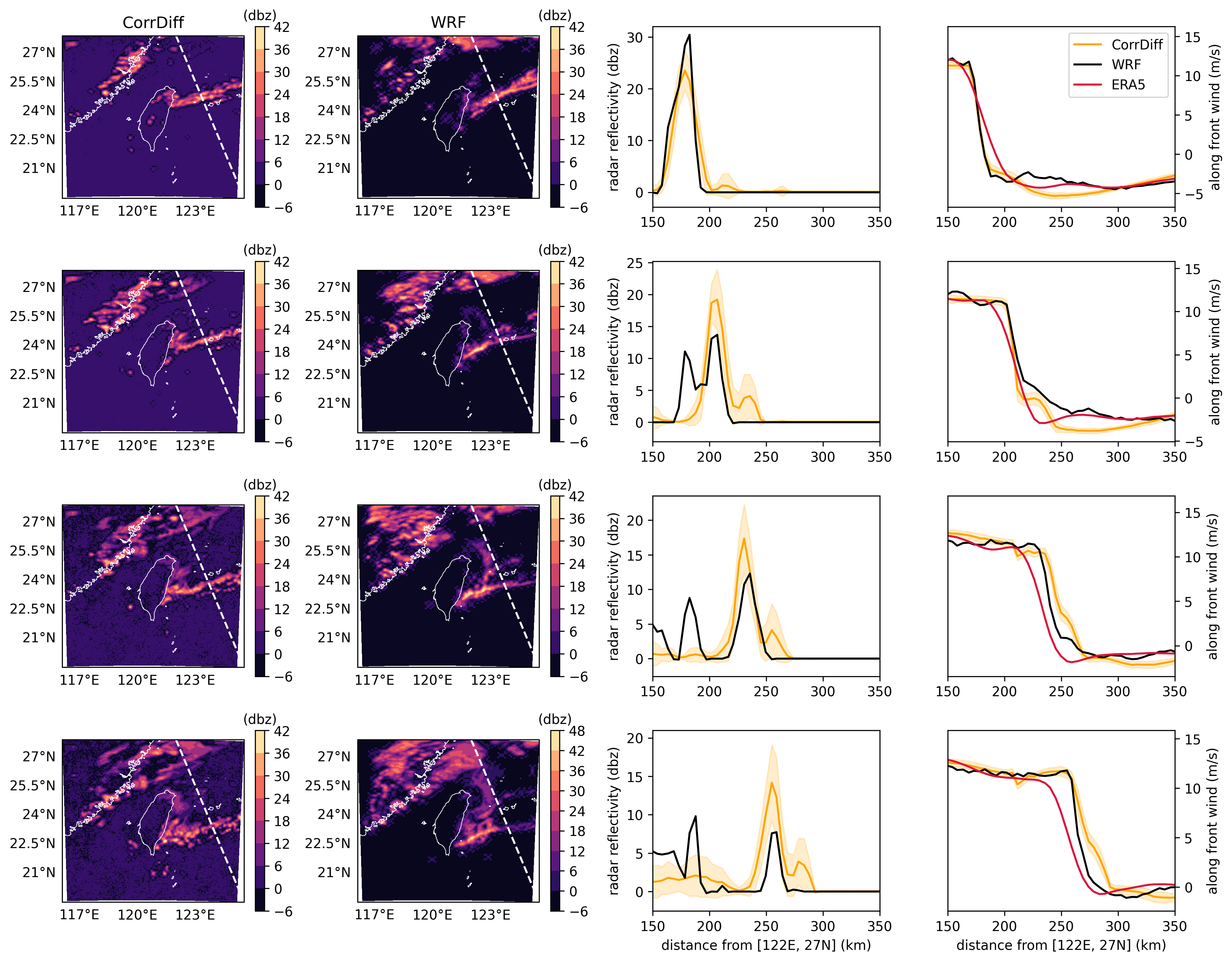} 
\caption{Same as \ref{fig:front_ref_2022} but for time:  2023-02-13 08:00:00, 2023-02-13 10:00:00, 2023-02-13 12:00:00 and 2023-02-13 14:00:00 UTC.}  \label{fig:front_ref_2023}
\end{figure}

Building on the analysis of the front in the main text (see Fig. \ref{fig:generation-front}), we further examine the coherence of synthesized reflectivity in two frontal events: one from 2022 (mentioned above) and another from 2023. 
Figures \ref{fig:front_ref_2022} and \ref{fig:front_ref_2023} compare the reflectivity maps for the target and CorrDiff along with the cross-section of the along-front winds. 

This analysis shows that although CorrDiff does not always sharpen the fronts to meet the target data, it syntheses the radar reflectivity consistently with the other variables and respects the frontal boundary by keeping the warm sector (of negative along front winds) cloud-free. Here we choose to indicate the frontal boundary with the along front wind. In all frontal events we examined, except the 2022 event in the main paper, fronts are moving fast and have large scale components of across front temperature gradient, and winds, which complicates tracking them in these channels.

\subsection{Additional Typhoon Analysis}
\label{section:additional_typhoons}
As additional analysis we first analyze two out-of-sample typhoons and then analyze a large number of typhoons without CWA target data by comparing their diagnosed properties to observed records. 

Figures \ref{fig:Chanthu} and \ref{fig:Haikui} below show analyses of typhoons Chanthu (2021) and Haikui (2023) respectively. The former is in our out-of-sample year and the latter was received in an additional data from CWA specifically for this analysis. The performance of ERA5, and consequently CorrDiff, differs substantially between these two typhoons. Chanthu (2021) presents an exceptional challenge for low resolution models like the one underlying ERA5, due to its meridional trajectory along Taiwan's coast at a distance of about 100-200 km. In such a trajectory the impact of Taiwan 's steep topography (up to 3km elevation) depends on the size of the simulated typhoon. The small Chanthu (2021) simulated by the WRF model, with a radius of maximum winds of about 25km, is effectively over ocean. Conversely, the larger Chanthu (2021) in ERA5, with radius of maximum winds of about 100km, is significantly affected by the topography. 

As a result, the ERA5 Chanthu (2021) can have less than a third of the intensity its WRF counterpart, see the first and third columns in \ref{fig:Chanthu}. It often does not have a closed windspeed contour above $10 m s^{-1}$, while the WRF Chanthu (2021) displays a coherent structure at $50m s^{-1}$ contours. Such a gap between the target (WRF) and the condition (ERA5) is hard for CorrDiff to overcome, and indeed the model fails to recover the intensity of the typhoon for most days. Only on 2023-09-03 12:00:00 UTC (top row of \ref{fig:Chanthu}), when the typhoon has passed the island, does CorrDiff provide an improvement over ERA5.

\begin{figure}
    \centering        \hspace{0mm}\includegraphics[width=1\textwidth,clip]{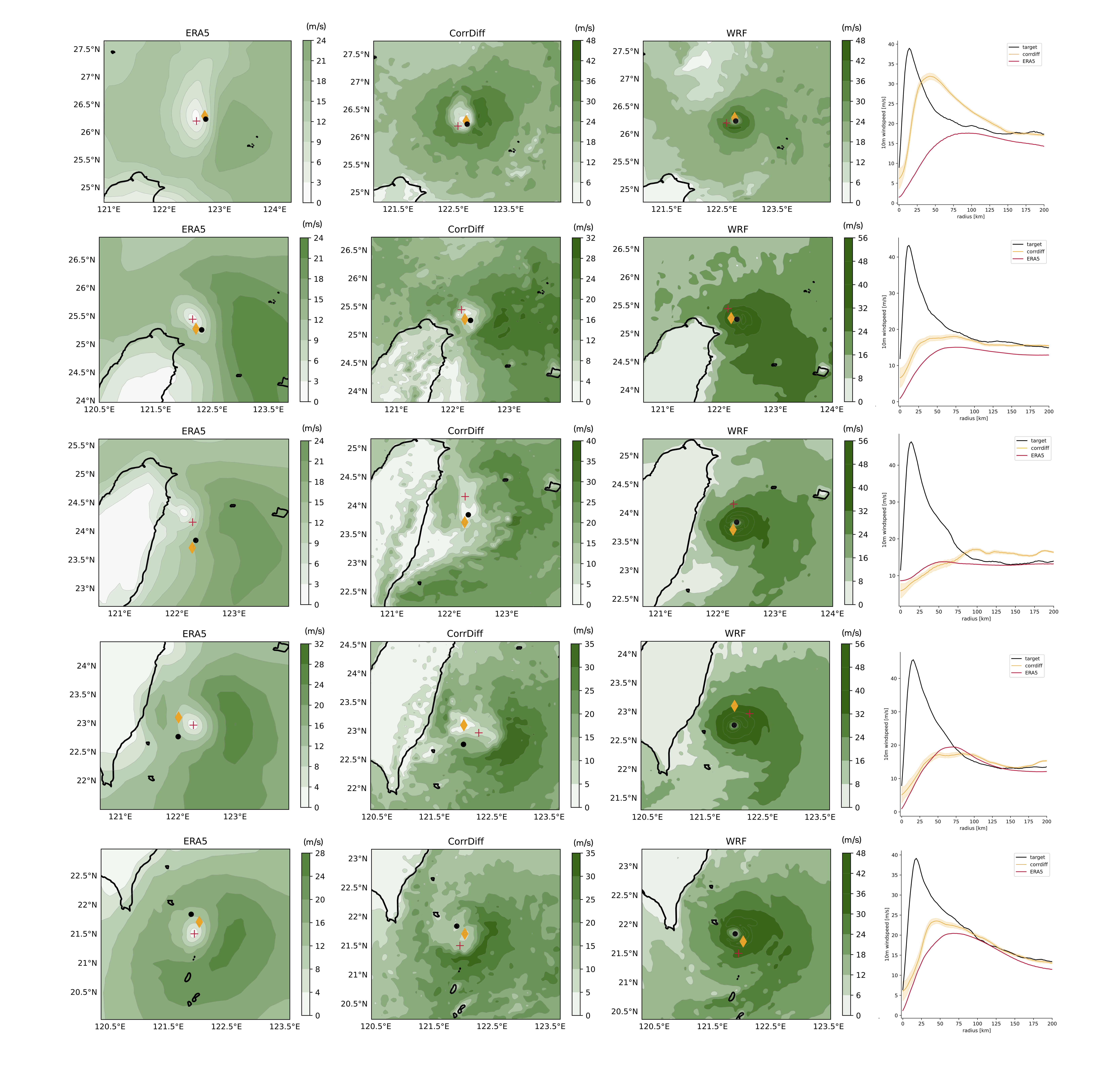} 
\caption{Windspeed maps and axisymmetric windspeed of typhoon Chanthu (2021). Left to right: windspeed in ERA5, CorrDiff, target (WRF model) and their axisymmetric profiles. The red `+`, orange diamond and black dot show the storm center for the ERA5, CorrDiff, WRF respectively. Time is increasing from bottom to top: 2021-09-02 12:00:00 UTC, 2021-09-02 18:00:00 UTC, 2021-09-03 00:00:00 UTC, 2021-09-03 06:00:00 UTC, 2021-09-03 12:00:00.}  \label{fig:Chanthu}
\end{figure}

Unlike Chanthu (2021), typhoon  Haikui (2023) has a zonal trajectory leading to landfall in southern Taiwan. CorrDiff improves upon ERA5 by correcting about 50\% of the intensity error and contracts the typhoon's radius of maximum winds.

\begin{figure}
    \centering        \hspace{0mm}\includegraphics[width=1\textwidth,clip]{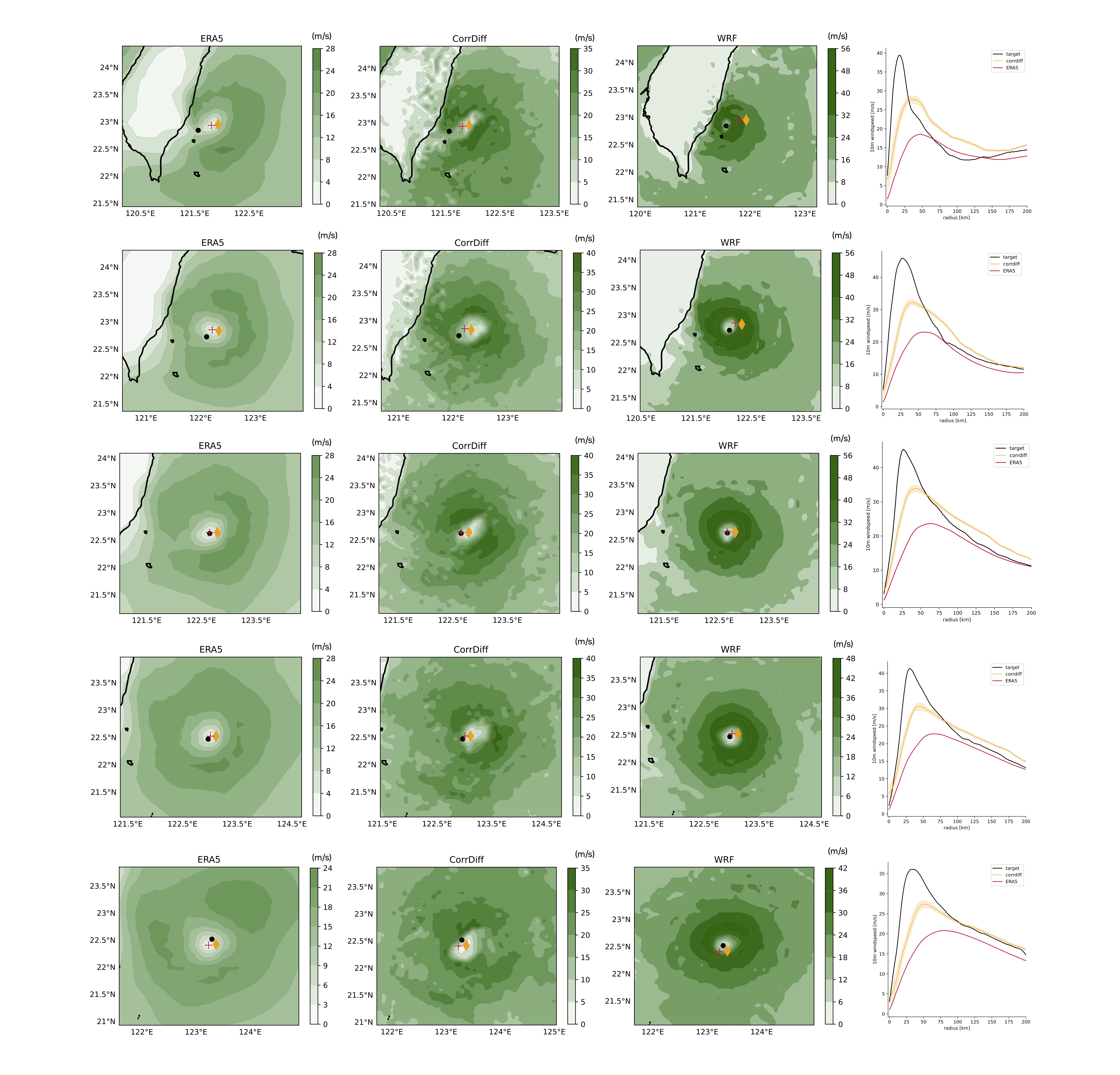} 
\caption{Same as \ref{fig:Chanthu} but for typhoon Haikui (2023). Time is increasing from bottom to top: 2023-09-02 18:00:00 UTC, 2023-09-02 21:00:00 UTC, 2023-09-03 00:00:00 UTC, 2023-09-03 03:00:00 UTC, 2023-09-03 06:00:00 UTC.}  \label{fig:Haikui}
\end{figure}

\begin{figure}
    \centering        \hspace{0mm}\includegraphics[width=1\textwidth,clip]{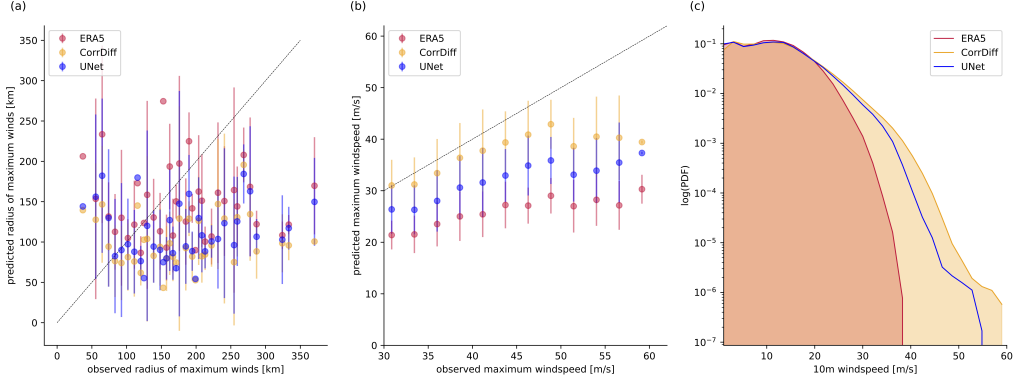} 
\caption{Comparison of predicted vs. observed (a) radius (km) and (b) value ($m s^{-1}$) of maximum axisymmetric windspeed ($m s^{-1}$) for all typhoons found in the domain with windspeed of $30 m s^{-1}$ or more spanning the years 1980-2020. Predictions are compared via the mean (dot) and one standard deviation (bars) of all predictions falling within each observational bin of the reference JMA observations. Panel (c) shows the log(PDF) of the predicted windspeed flattened across all of these typhoon instances.}  \label{fig:SI4}
\end{figure}

Further analysis compares CorrDiff-simulated typhoons in the CWA region with historical records to evaluate typhoons for which no target data exists.  The Japan Meteorological Agency best track data (JMA tracks) \cite{barcikowska2012usability} includes the maximum windspeed (intensity) and radius of maximum windspeed (size) of typhoons in the West Pacific for several decades.
We identified 648 instances of typhoons with intensities of $30 m s^{-1}$ or greater within the CWA domain from 1980 to 2020. Panels (a) and (b) of Fig. \ref{fig:SI4} display the storm size and intensity, respectively, revealing the expected correction for ERA5 typhoons achieved through the application of CorrDiff downscaling. One limitation of CorrDiff is that it reduces the size of all storms, including those with the correct size or those already too small in the ERA5 input data (panel a). The main benefit is improved windspeeds, removing most of the error between the ERA5 and the observed records for windspeeds up to $50 m s^{-1}$ (panel b); though stronger storms have room for improvement. By correcting ERA5 toward JMA tracks, CorrDiff generates a five-fold increase in the probability of windspeed values exceeding hurricane-force winds (i.e., $33 m s^{-1}$), see panel c. To the extent that JMA tracks can serve as ground truth, such distribution shift has significant societal implications, as these low-probability, high-impact events represent a substantial portion of the overall risk.

The UNet alone presents a less attractive alternative for downscaling typhoons, as its maximum axisymmetric windspeed is consistently positioned between the ERA5 and the CorrDiff values. CorrDiff offers a meaningful improvement over the UNet in both the maximum of the axisymmetric wind speed (panel b) and the wind speed maxima (panel c).

\subsection{Energy efficiency and latency of CorrDiff downscaling inference compared to WRF simulations}
\label{sec:speedup}
Comparing the performance of statistical downscaling like CorrDiff with dynamical downscaling like the WRF-CWA is  challenging due to their different approaches and outcomes. 
Statistical downscaling produces a high-resolution state for a subset of channels at time $t$ from a different (potentially larger) set of channels at lower resolution at the same time. Dynamical downscaling produces a full, high-resolution, state vector at time $t+\Delta t$ from both high and low resolution state vector at $t$ using a numerical time-stepper. The time step ($\Delta t$) here is constrained by numerical stability and can be of the order of seconds. Thus, to produce the prediction at +1h, the dynamical downscaling model might run hundreds of autoregressive steps while the statistical downscaling will make a single inference.

Nonetheless, from a utility perspective, both approaches are used for producing a high resolution state at a given time. Thus, we can compare the latency and energy required to obtain a single high resolution prediction in Taiwan from the two approaches. We compare the speed of CorrDiff against the operational WRF run by CWA on their respective hardware. 

The CWA-WRF is run on Fujitsu FX-100 system, with each node equipped with 32 SPARC64 Xifx CPU cores. A 13-hour deterministic CWA-WRF forecast (excluding data assimilation) is run on 928 CPU cores (across 29 nodes with a maximum system memory of 6.9GB per node) and takes about 20 minutes. CorrDiff inference is run on a single NVIDIA H100 Generation GPU, which takes 0.18 sec per downscaling sample. Given a global model 1-hour lead time forecast, the CorrDiff statistical downscaling on a single GPU is about $500$ times faster than the dynamic downscaling that runs CWA-WRF on 928 CPUs. Moreover, CorrDiff is about $10,000$ times more energy efficient. Since the individual samples are computed independently, conditioned on given global model data (which both systems need), CorrDiff can be run for the 13 hours on 13 GPUs, thus obtaining about a 13x speedup for the 13-hour forecast over the above results (but with the same energy efficiency).  These results ignore the additional compute needed for the regional DA in CWA-WRF, which is absent in CorrDiff (and likely impacts its performance). The regional DA in CWA, depending on the method used, can increase the compute for the CWA-WRF by a factor of 1.5 or 2 \cite{chen2020improving}.

\begin{table}[h]
    \centering
    \begin{tabular}{lcccc}
        \toprule
        & Hardware & Latency (sec/FH) & Power (J/sec) & Energy (kJ/FH) \\
        \midrule
        WRF-CWA & 928 CPUs & 91.38 & 15.15 & 1285.46 \\
        CorrDiff & 1 GPU & 0.18 & 700 & 0.126 \\
        \bottomrule
    \end{tabular}
    \caption{A comparison of running the WRF model on the CWA system with CorrDiff inference on a single NVIDIA H100 GPU. Latency is given per Forecast Hour (FH) and Power is given in Joule/sec (W) per a single hardware unit (a CPU or a GPU), while Energy is for the entire forecast system (928 CPU for WRF-CWA) per FH.}
    \label{tab:your_label}
\end{table}
\subsection{Statistical significance of the CRPS metrics \label{sec:significance}}
\begin{figure}
    \centering
    \includegraphics{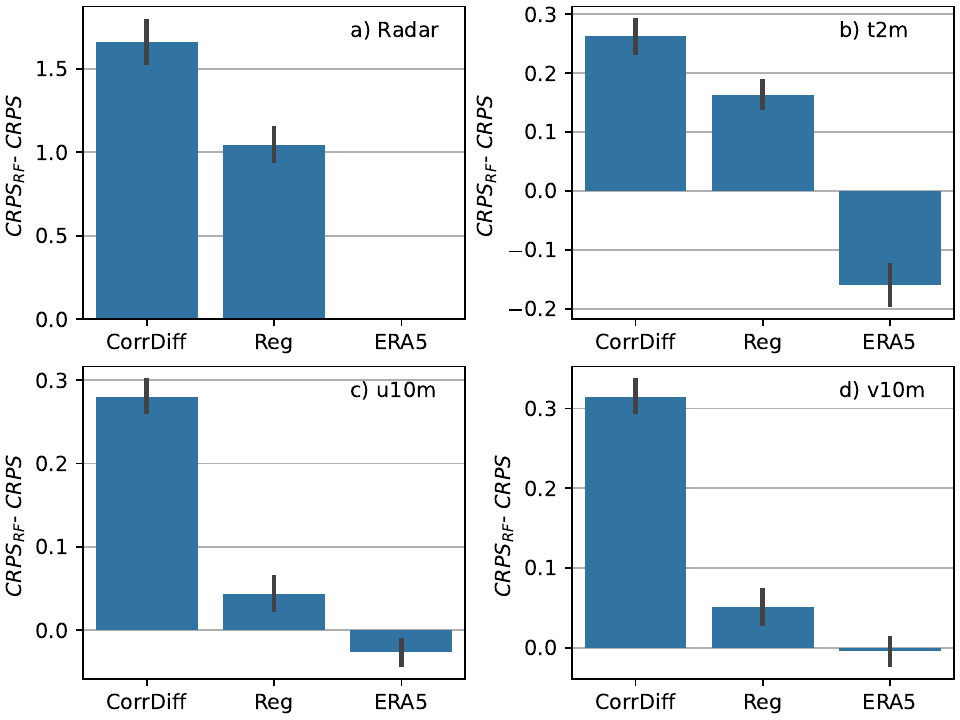}
    \caption{Boostrapping analysis of skill improvements presented in Table \ref{tab:big-score}. The error bar shows a 95\% confidence interval for the mean obtained by bootstrapping. }
    \label{fig:metric-error-bar}
\end{figure}

We further diagnose the statistical significance of the metrics presented in Table \ref{tab:big-score}. We show this significance both graphically and using hypothesis testing.

By evaluating our models on the same set of times, we can take advantage of paired statistical tests, which provide much more power. Given the scores for two models $x_i$ and $y_i$ evaluated over times $i$, we find that $Var(x_i - y_i) \ll Var(x_i)$. So even though the scores may vary in time, the improvements are robust. Figure \ref{fig:metric-error-bar} shows the improvement in CRPS relative to the RF baseline. In all cases, the improvements of our models (CorrDiff and UNet) are larger than the error bars.

We further elaborate on the significance with formal hypothesis testing for the radar field. The standard non-parametric test for assessing whether $x_i > y_i$ is the Wilcoxon signed-rank test for $d_i = (x_i - y_i)$. An alternative test is the binomial test of $x_i > y_i$. In both cases, the p-values comparing the RF baseline with CorrDiff are smaller than $10^{-30}$. This remarkably low p-value may seem surprising, but CorrDiff has lower CRPS in 205 out of 205 times. The likelihood this occurring by random chance is vanishingly small.
\end{document}